\documentclass{article}

\PassOptionsToPackage{numbers, compress}{natbib}
\usepackage[final]{neurips_2020}
\usepackage[utf8]{inputenc} 
\usepackage[T1]{fontenc}    
\usepackage{hyperref}       
\usepackage{url}            
\usepackage{booktabs}       
\usepackage{amsfonts}       
\usepackage{nicefrac}       
\usepackage{microtype}      
\usepackage{graphicx}
\usepackage{amsmath}
\usepackage{algorithm}
\usepackage{algorithmic}
\usepackage{xcolor}
\usepackage{multirow}
\usepackage{subcaption}
\usepackage{wrapfig}

\newcommand{\cut}[1]{}
\newcommand{\keypoint}[1]{\noindent\textbf{#1}\quad}

\DeclareMathOperator*{\argmin}{arg\,min}

\title{Flexible Dataset Distillation:\\ Learn Labels Instead of Images}

\author{%
  Ondrej Bohdal, Yongxin Yang, Timothy Hospedales \\
  School of Informatics \\
  The University of Edinburgh \\
  \texttt{\{ondrej.bohdal, yongxin.yang, t.hospedales\}@ed.ac.uk} \\
}

\begin{document}

\maketitle

\begin{abstract}
We study the problem of dataset distillation -- creating a small set of synthetic examples capable of training a good model. In particular, we study the problem of \emph{label} distillation -- creating synthetic labels for a small set of real images, and show it to be more effective than the prior \emph{image}-based approach to dataset distillation. Methodologically, we introduce a more robust and flexible meta-learning algorithm for distillation, as well as an effective first-order strategy based on convex optimization layers. Distilling labels with our new algorithm leads to improved results over prior image-based distillation. More importantly, it leads to clear improvements in flexibility of the distilled dataset in terms of compatibility with off-the-shelf optimizers and diverse neural architectures. Interestingly, label distillation can be applied across datasets, for example enabling learning Japanese character recognition by training only on synthetically labeled English letters.
\end{abstract}

\section{Introduction}
Distillation is a topical area of neural network research that initially began with the goal of extracting the knowledge of a large pre-trained model and compiling it into a smaller model, while retaining similar performance \cite{Hinton2014DistillingNetwork}. The notion of distillation has since found numerous applications and uses including the possibility of \emph{dataset} distillation \cite{Wang2018DatasetDistillation}: extracting the knowledge of a large dataset and compiling it into a small set of carefully crafted examples, such that a model trained on the small dataset alone achieves good performance. This is of scientific interest as a tool to study neural network generalisation under small sample conditions. More practically, it has the potential to address the large and growing logistical and energy hurdle of neural network training, if adequate neural networks can be quickly trained on small distilled datasets rather than massive raw datasets.

Nevertheless, progress towards the vision of dataset distillation has been limited as the performance of existing methods \citep{Wang2018DatasetDistillation, Sucholutsky2019Soft-LabelDistillation} trained from random initialization is far from that of full dataset supervised learning. More fundamentally, existing approaches are relatively \emph{inflexible} in terms of the distilled data being over-fitted to the training conditions under which it was generated. While there is some robustness to the choice of initialization weights \cite{Wang2018DatasetDistillation}, the distilled dataset is largely specific to the architecture used to train it (preventing its use to accelerate neural architecture search, for example), and must use a highly-customized learner (a specific image visitation sequence, a specific sequence of carefully chosen meta-learned learning rates, and a specific number of learning steps). Altogether these constraints mean that existing distilled datasets are not general purpose enough to be useful in practice, e.g. with off-the-shelf learning algorithms. We propose a more \emph{flexible} approach to dataset distillation underpinned by both algorithmic improvements and changes to the problem definition.

\begin{figure}
  \centering
   \includegraphics[width=0.7\columnwidth]{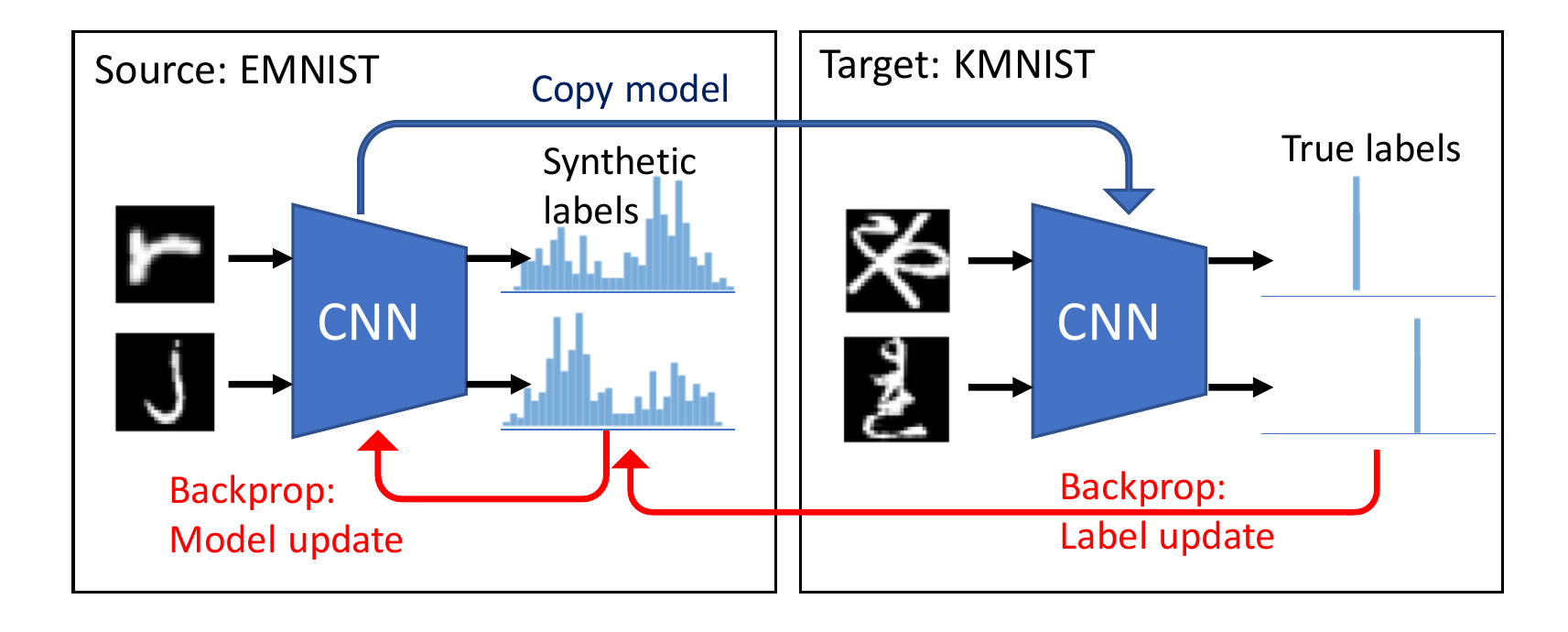}
  \caption{Label distillation enables training a model that can classify Japanese characters after being trained only on English letters and synthetic labels. Labels are updated only during meta-training, after which a new model is trained, using only the few source examples and their synthetic labels.}
  \label{fig:LabelDistillationPipeline}
\end{figure}

Rather than creating synthetic images \citep{Wang2018DatasetDistillation} for arbitrary labels, or a combination of synthetic images and soft labels \citep{Sucholutsky2019Soft-LabelDistillation}, we focus on crafting synthetic labels for arbitrarily chosen standard images. Compared to these prior approaches focused on synthetic images, label distillation benefits from exploiting the data statistics of natural images and the lower-dimensionality of labels compared to images as parameters for meta-learning. Practically, this leads to improved performance compared to prior image distillation approaches. As a byproduct, this enables a new kind of cross-dataset knowledge distillation (Figure~\ref{fig:LabelDistillationPipeline}). One can learn solely on a source dataset (such as English characters) with synthetic distilled labels, and apply the learned model to recognise concepts in a disjoint target dataset (such as Japanese characters). Surprisingly, it turns out that models can make progress on learning to recognise Japanese only through exposure to English characters with synthetic labels.

Methodologically, we define a new meta-learning algorithm for distillation that does not require costly evaluation of multiple inner-loop (model-training) steps for each iteration of distillation. More importantly our algorithm leads to a more flexible distilled dataset that is better transferable across optimizers, architectures, learning iterations, etc. Furthermore, where existing dataset distillation algorithms rely on second-order gradients, we introduce an alternative learning strategy based on convex optimization layers that avoids high-order gradients and provides better optimization, thus improving the quality of the distilled dataset.

In summary, we contribute: (1) A dataset distillation method that produces flexible distilled datasets that exhibit transferability across learning algorithms. This brings us one step closer to producing useful general-purpose distilled datasets. (2) Our distilled datasets can be used to train higher performance models than those prior work. (3) We introduce the novel concept of cross-dataset distillation, and demonstrate proofs of concept, such as English$\to$Japanese letter recognition.

\section{Related work}
\keypoint{Dataset distillation}
Most closely related to our work is Dataset \citep{Wang2018DatasetDistillation} and Soft-Label Dataset Distillation \citep{Sucholutsky2019Soft-LabelDistillation}. They focus on distilling a dataset or model \citep{Micaelli2019Zero-shotMatching} into a small number of example images, which are then used to train a new model. This can be seen as solving a meta-learning problem with respect to model's training data \citep{Hospedales2020Meta-LearningSurvey}. The common approach is to initialise the distilled dataset randomly, use the distilled data to train a model, and then backpropagate through the model and its training updates to take gradient steps on the dataset. Since the `inner' model training algorithm is gradient-based, this leads to high-order gradients. To make this process tractable, the original Dataset Distillation  \citep{Wang2018DatasetDistillation} uses only a few gradient steps in its inner loop (as per other famous meta-learners \cite{Finn2017Model-AgnosticNetworks}). To ensure that sufficient learning progress is made with few updates, it also meta-learns a fixed set of optimal learning rates to use at each step. This balances tractability \& efficacy, but causes the distilled dataset to be `locked in' to the customized optimizer rather than serve as a general purpose dataset, which also prevents its use for NAS \citep{Shleifer2019ProxyNetworks}. In this work we define an online meta-learning procedure that simultaneously learns the dataset and the base model. This enables us to tractably take more gradient steps and ultimately produce a performant yet flexible general purpose distilled dataset.

There are various motivations for dataset distillation, but the most practically intriguing is to summarize a dataset in a compressed form to accelerate model training. In this sense it is related to dataset pruning \citep{Angelova2005PruningCategories, Felzenszwalb2010ObjectModels, Lapedriza2013AreValuable}, core-set construction \citep{Tsang2005CoreSets, Bachem2017PracticalLearning, Sener2018ActiveApproach} and instance selection \citep{Olvera-Lopez2010AMethods} focusing on dataset summarization through a small number of examples. Summarization methods \emph{select} a relatively large part of the data (e.g. at least 10\%), while distillation extends down to using 10 images per category ($\approx0.2\%$ of CIFAR-10 data) through example \emph{synthesis}. We keep original data (like summarization methods), but synthesize labels (like distillation). This leads to a surprising observation -- it is possible to synthesize labels for a few fixed examples so a model trained on these examples can directly (without any fine-tuning) solve a different problem with a different label space (Figure~\ref{fig:LabelDistillationPipeline}).

\keypoint{Meta-learning}
Meta-learning algorithms can often be grouped \cite{Hospedales2020Meta-LearningSurvey} into offline approaches (e.g. \cite{Wang2018DatasetDistillation,Finn2017Model-AgnosticNetworks}) that do inner optimization at each step of outer optimization; and online approaches that solve the base and meta-learning problem simultaneously (e.g. \cite{Balaji2018MetaReg:Meta-Regularization,Li2019Feature-CriticGeneralization}). Meta-learning relates to hyperparameter optimization, for example \cite{Maclaurin2015Gradient-basedLearning, Lorraine2020OptimizingDifferentiation} efficiently unroll through many steps of optimization like offline meta-learning, while \cite{Luketina2016ScalableHyperparameters} optimize hyperparameters and the base model like online meta-learning. Online approaches are typically faster, but optimize meta-parameters for a single problem. Offline approaches are slower and typically limit the length of the inner optimization for tractability, but can often find meta-parameters that solve a distribution of tasks (as different tasks are drawn in each outer-loop iteration). In dataset distillation, the notion of `distribution over tasks' corresponds to finding a dataset that can successfully train a network in many settings, such as different initial conditions \cite{Wang2018DatasetDistillation}.
Our distillation algorithm is a novel hybrid of these two families. We efficiently solve the base and meta-tasks simultaneously like online approaches, and so are able to use more inner-loop steps. However, we also learn to solve many `tasks' by detecting meta-overfitting and sampling a new `task' when this occurs. This leads to a great combination of efficacy and efficiency.

Finally, most gradient-based meta-learning algorithms rely on costly and often unstable higher-order gradients \cite{Hospedales2020Meta-LearningSurvey,Finn2017Model-AgnosticNetworks,Wang2018DatasetDistillation}, or else make simple shortest-path first-order approximations \cite{Nichol2018OnAlgorithms}. Instability and large variance of higher-order gradients may make meta-learning less effective \cite{Liu2019TamingLearning, Farquhar2019LoadedLearning}, so we found inspiration in recent approaches in few-shot learning \cite{Bertinetto2019Meta-LearningSolvers,Lee2019Meta-LearningOptimization} that avoid this issue through the use of convex optimization layers. We introduce the notion of a pseudo-gradient that enables this idea to scale beyond the few-shot setting to general meta-learning problems such as dataset distillation.

\section{Methods}
We aim to meta-learn soft synthetic labels for a small fixed set of real \textit{base} examples that can be used to train a randomly initialized model. This corresponds to an objective such as Eq.~\ref{eqn:bilevelopt}:
\begin{equation}
\label{eqn:bilevelopt}
    \tilde{\boldsymbol{Y}}_{\mathcal{S}}^{*}=\argmin_{\tilde{\boldsymbol{Y}}_{\mathcal{S}}} \sum_{\boldsymbol{x}, \boldsymbol{y} \sim \mathcal{T}} L \left(f_{\boldsymbol{\Theta}^\prime}\left(\boldsymbol{x}\right), \boldsymbol{y}\right), \quad \mathrm{with} \quad \boldsymbol{\Theta}^\prime = \boldsymbol{\Theta}-\alpha \nabla_{\boldsymbol{\Theta}} \sum_{\boldsymbol{\tilde{x}}, \boldsymbol{\tilde{y}} \sim \mathcal{S}} L \left(f_{\boldsymbol{\Theta}}\left(\boldsymbol{\tilde{x}}\right), \boldsymbol{\tilde{y}}\right),
\end{equation}
where $\tilde{\boldsymbol{Y}}_{\mathcal{S}} \in \mathbb{R}^{N\times C}$ are the distilled labels for $N$ base examples $\tilde{\boldsymbol{X}}_{\mathcal{S}} \in \mathbb{R}^{N\times D}$ (together forming synthetic set $\mathcal{S}$). Each image has dimensionality $D$, and there are $C$ target classes. Further, $\mathcal{T}$ is the target set with real training data, $f_{\boldsymbol{\Theta}}(\cdot)$ is a model with parameters $\boldsymbol{\Theta}$, $\alpha$ is the learning rate and $L$ is the cross-entropy loss. We assume the loss is twice-differentiable, which is true for most current machine learning models and problems. One gradient step is shown above, but in general there may be multiple steps. Cross-entropy loss for predicted soft labels $\boldsymbol{\hat{y}}=f_{\boldsymbol{\Theta}^\prime}\left(\boldsymbol{x}\right)$ and true one-hot labels $\boldsymbol{y}$ for an example $\boldsymbol{x}$ is defined as $L(\boldsymbol{\hat{y}}, \boldsymbol{y}) = - \sum_{c=1}^C y_c \log\left(\hat{y}_c\right)$ (similar for synthetic labels).



One option to achieve objective in Eq.~\ref{eqn:bilevelopt} would be to follow \citet{Wang2018DatasetDistillation} and simulate the whole training procedure for multiple gradient steps $\nabla_{\boldsymbol{\Theta}}$ within the inner loop. However, this requires back-propagating through a long inner loop, and ultimately requires a fixed training schedule with optimized learning rates for strong performance. We aim to produce a dataset that can be used in a standard training pipeline downstream (e.g. Adam optimizer with the default parameters).

Our first modification to the standard pipeline is to perform gradient descent iteratively on the model and the distilled labels, rather than performing many inner (model) steps for each outer (dataset) step. This increases efficiency significantly due to a shorter compute graph for backpropagation. Nevertheless, when there are very few training examples, the model converges quickly to an over-fitted local minimum, likely within a few hundred iterations. To manage this, our innovation is to detect overfitting when it occurs, reset the model to a new random initialization and keep training. Specifically, we measure the moving average of target problem accuracy, and when it has not improved for set number of iterations, we reset the model. This periodic reset of the model after varying number of iterations is helpful for learning labels that are useful for all stages of training and thus less sensitive to the number of iterations used. To ensure scalability to any number of examples, we sample a minibatch of base examples and synthetic labels and use those to update the model, which also better aligns with standard training practice. Once label distillation is done, we train a new model from scratch using random initial weights, given the base examples and learned synthetic labels. Our algorithm returns a reasonable number of training steps $T_i$, which allows us to stop training the new model early and prevent over-fitting to the small synthetic dataset. 

We propose and evaluate two label distillation algorithms: a second-order version that performs one update step within the inner loop; and a first-order version that uses a closed form solution of ridge regression to find optimal classifier weights for the base examples.

\keypoint{Vanilla second-order version} The training includes both inner and outer loop. The inner loop consists of one update of the model parameters $\boldsymbol{\Theta}^\prime = \boldsymbol{\Theta}-\alpha \nabla_{\boldsymbol{\Theta}} \sum_{\boldsymbol{\tilde{x}}, \boldsymbol{\tilde{y}} \sim \mathcal{S}} L \left(f_{\boldsymbol{\Theta}}\left(\boldsymbol{\tilde{x}}\right), \boldsymbol{\tilde{y}}\right)$, through which we then backpropagate to update the synthetic labels $\boldsymbol{\Tilde{Y}}_{\mathcal{S}} \leftarrow \boldsymbol{\Tilde{Y}}_{\mathcal{S}} - \beta\nabla_{\boldsymbol{\Tilde{Y}}_{\mathcal{S}}} \sum_{\boldsymbol{x}, \boldsymbol{y} \sim \mathcal{T}} L \left(f_{\boldsymbol{\Theta}^\prime}\left(\boldsymbol{x}\right), \boldsymbol{y}\right)$ (same notation as in Eq. \ref{eqn:bilevelopt}). We do a standard update of the model parameters $\boldsymbol{\Theta}$ using outer-loop learning rate $\beta$ after updating the synthetic labels, which could in theory be combined with the inner-loop update. The method is otherwise similar to our first-order ridge regression method, which is summarized in Algorithm \ref{alg:labeldistillationmetaopt}. Real training set examples are used when updating the synthetic labels $\boldsymbol{\Tilde{Y}}_{\mathcal{S}}$, but for updating the model $\boldsymbol{\Theta}$ we only use the synthetic labels and the base examples. After each update of the labels, we normalize them to represent a valid probability distribution. This makes them interpretable and has led to improvements compared to unnormalized labels.



\keypoint{Intuition}
We analysed how the synthetic labels are meta-learned by the second-order algorithm for a simple one-layer model $\boldsymbol{\theta}$, with a sigmoid output unit $\sigma$ and two classes (details are in the supplementary). Considering one example at a time for simplicity with $\left(\boldsymbol{x}, y\right) \sim \mathcal{T}$ and $\left(\boldsymbol{\tilde{x}}, \tilde{y}\right) \sim \mathcal{S}$, the meta-gradient is $\nabla_{\tilde{y}} L\left(\sigma(\boldsymbol{\theta}^{\prime T}\boldsymbol{x}), y\right) = \alpha\left(\sigma\left(\boldsymbol{\theta}^{\prime T}\boldsymbol{x}\right)-y\right)\boldsymbol{x}^T\boldsymbol{\Tilde{x}}.$ This shows the synthetic label is updated proportionally to the similarity of the training set example $\boldsymbol{x}$ and base example $\boldsymbol{\Tilde{x}}$ as well as the difference between the prediction and the true real training set label. Thus synthetic labels capture the different degrees of similarity between a base example and examples from different classes in the target dataset. For example, in cross-dataset, the KMNIST `{\tt Ya}' \includegraphics[height=1.4\fontcharht\font`\B]{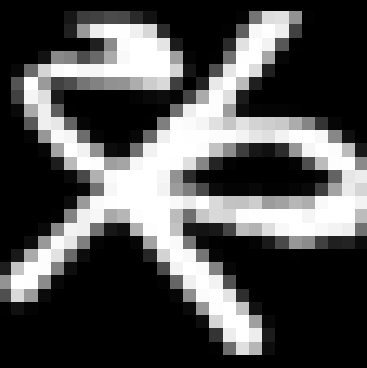} has no corresponding English symbol, but could be learned by partially assigning its label to similar looking English `{\tt X}'s and `{\tt R}'s.

\begin{algorithm*}[t]
   \caption{Label distillation with ridge regression (RR)}
   \label{alg:labeldistillationmetaopt}
\begin{algorithmic}[1]
    \STATE {\bfseries Input:} $\mathcal{S}$: synthetic set with $N$ initially unlabelled base examples $\tilde{\boldsymbol{X}}_{\mathcal{S}}$; $\mathcal{T}$: labelled target set with real training examples; $\beta$: step size; $\alpha$: pseudo-gradient step size; $N_o, N_i$: outer (resp. inner) loop batch size; $C$: number of classes in the target set; $\lambda$: RR regularization parameter
    \STATE {\bfseries Output:} distilled labels $\tilde{\boldsymbol{Y}}_{\mathcal{S}}$ and a reasonable number of training steps $T_i$
    \STATE $\tilde{\boldsymbol{Y}}_{\mathcal{S}}\leftarrow \boldsymbol{1}/C$\hfill //Uniformly initialize synthetic labels
    \STATE $\boldsymbol{\Theta}\sim p(\boldsymbol{\Theta})$ \hfill //Randomly initialize feature extractor parameters
    \STATE $\boldsymbol{W} \leftarrow \boldsymbol{0}$ \hfill //Initialize global RR classifier weights
    \WHILE{$\tilde{\boldsymbol{Y}}_{\mathcal{S}}$ not converged}
    \STATE $\left(\tilde{\boldsymbol{X}}, \tilde{\boldsymbol{Y}}\right) \sim \mathcal{S} =\left(\boldsymbol{\Tilde{X}}_\mathcal{S}, \boldsymbol{\Tilde{Y}}_\mathcal{S}\right) $\hfill//Sample a minibatch of $N_i$ synthetic examples 
    \STATE $\left(\boldsymbol{X}, \boldsymbol{Y}\right) \sim \mathcal{T}$\hfill//Sample a minibatch of $N_o$ real training examples
    \STATE Calculate $\boldsymbol{W}_l$ using Eq. \ref{eqn:rrcalc}\hfill //Calculate current minibatch RR classifier weights
    \STATE $\boldsymbol{W} \leftarrow (1-\alpha) \boldsymbol{W} + \alpha \boldsymbol{W}_l$\hfill //Update global RR classifier weights
    \STATE $\boldsymbol{\Tilde{Y}}_{\mathcal{S}} \leftarrow \boldsymbol{\Tilde{Y}}_{\mathcal{S}} - \beta\nabla_{\boldsymbol{\Tilde{Y}}_{\mathcal{S}}} \sum_{\left(\boldsymbol{x}, \boldsymbol{y}\right) \in \left(\boldsymbol{X}, \boldsymbol{Y}\right)} L \left(f_{\boldsymbol{W}\circ\boldsymbol{\Theta}}\left(\boldsymbol{x}\right), \boldsymbol{y}\right)$\hfill//Update synthetic labels
    \STATE $\boldsymbol{\Theta} \leftarrow \boldsymbol{\Theta}-\beta \nabla_{\boldsymbol{\Theta}} \sum_{\left(\boldsymbol{\Tilde{x}}, \boldsymbol{\Tilde{y}}\right) \in \left(\boldsymbol{\Tilde{X}}, \boldsymbol{\Tilde{Y}}\right)} L \left(f_{\boldsymbol{W} \circ \boldsymbol{\Theta}}\left(\boldsymbol{\tilde{x}}\right), \boldsymbol{\tilde{y}}\right)$\hfill//Update feature extractor
    \IF{$\mathbb{E}_{\boldsymbol{x}, \boldsymbol{y} \sim \mathcal{T}} \left[L \left( f_{\boldsymbol{W}\circ\boldsymbol{\Theta}}\left(\boldsymbol{x}\right), \boldsymbol{y}\right)\right]$ did not improve} 
    \STATE $\boldsymbol{\Theta}\sim p(\boldsymbol{\Theta})$\hfill//Reset feature extractor
    \STATE $\boldsymbol{W} \leftarrow\boldsymbol{0}$ \hfill//Reset global RR classifier weights
    \STATE $T_i\leftarrow$ iterations since previous reset \hfill //Record time to overfit
    \ENDIF
    \ENDWHILE
\end{algorithmic}
\end{algorithm*}

\keypoint{First-order version with ridge regression}
To avoid second-order gradients, we propose a first-order version that uses pseudo-gradient generated via a closed form solution to ridge regression (RR). We use a RR layer as the final output layer of our base network -- we decompose the original model $\boldsymbol{\Theta}$ used for the second-order version into feature extractor $\boldsymbol{\Theta}$ and RR classifier $\boldsymbol{W}$. RR layers have previously been used for few-shot learning \citep{Bertinetto2019Meta-LearningSolvers} \emph{within} minibatch. We extend it to learn global weights that persist across minibatches. Local ridge regression problem can be defined and solved as:
\begin{equation}
\begin{aligned}
\boldsymbol{W}_l &=\argmin_{\boldsymbol{W}^\prime} \left\|\boldsymbol{\Tilde{Z}} \boldsymbol{W}^\prime-\boldsymbol{\Tilde{Y}}\right\|^{2}+\lambda\left\|\boldsymbol{W}^\prime\right\|^{2} \\
&=\left(\boldsymbol{\Tilde{Z}}^{T} \boldsymbol{\Tilde{Z}}+\lambda \boldsymbol{I}\right)^{-1} \boldsymbol{\Tilde{Z}}^{T} \boldsymbol{\Tilde{Y}},
\end{aligned}
\end{equation}
where $\boldsymbol{\Tilde{Z}}=f_{\boldsymbol{\Theta}}\left(\boldsymbol{\Tilde{X}}\right)$ are the input embeddings for a minibatch of $N_i$ base examples and synthetic labels $\left(\boldsymbol{\Tilde{X}}, \boldsymbol{\Tilde{Y}}\right) \sim \left(\boldsymbol{\Tilde{X}}_\mathcal{S}, \boldsymbol{\Tilde{Y}}_\mathcal{S}\right)$. $\boldsymbol{W}_l$ represents the ridge regression weights, $\boldsymbol{I}$ is the identity matrix and $\lambda$ is the regularization parameter. Following \citet{Bertinetto2019Meta-LearningSolvers}, we use Woodbury formula \citep{Petersen2012TheCookbook}: 
\begin{equation}
\label{eqn:rrcalc}
    \boldsymbol{W}_l=\boldsymbol{\Tilde{Z}}^{T}\left(\boldsymbol{\Tilde{Z}} \boldsymbol{\Tilde{Z}}^{T}+\lambda \boldsymbol{I}\right)^{-1} \boldsymbol{\Tilde{Y}},
\end{equation}
which allows us to use matrix $\boldsymbol{\Tilde{Z}} \boldsymbol{\Tilde{Z}}^{T}$ with dimensionality depending on the square of the number of inputs (minibatch size) rather than the square of the input embedding size. This makes the matrix inversion significantly less costly in practice. While ridge regression is usually oriented at regression problems, it has been shown \citep{Bertinetto2019Meta-LearningSolvers} to work well for classification when regressing label vectors.


\keypoint{Ridge regression with pseudo-gradients}
RR solves for the optimal local weights $\boldsymbol{W}_l$ that classify the features of the current minibatch examples. We exploit this local minibatch solution by taking a pseudo-gradient step that updates global weights $\boldsymbol{W}$ as $\boldsymbol{W} \leftarrow (1-\alpha) \boldsymbol{W} + \alpha \boldsymbol{W}_l$, with $\alpha$ being the pseudo-gradient step size. We can understand this as a pseudo-gradient as it corresponds to the step $\boldsymbol{W} \leftarrow \boldsymbol{W} - \alpha(\boldsymbol{W}-\boldsymbol{W}_l)$. We can then update the synthetic labels by back-propagating through local weights $\boldsymbol{W}_l$. Subsequent feature extractor updates on $\boldsymbol{\Theta}$ avoid second-order gradients. The process is summarised in Algorithm~\ref{alg:labeldistillationmetaopt}. 

\section{Experiments}
We perform two main types of experiments: (1) within-dataset distillation, when the base examples come from the target dataset and (2) cross-dataset distillation, when the base examples come from a different but related dataset. The dataset should be related because if there is a large shift in the domain (e.g. from characters to photos), then the feature extractor trained on the base examples would generalize poorly to the target dataset. We use MNIST, CIFAR-10 and CIFAR-100 for the task of within-dataset distillation, while for cross-dataset distillation we use EMNIST (``English letters''), KMNIST, Kuzushiji-49 (both ``Japanese letters''), MNIST (digits), CUB (birds) and CIFAR-10 (general objects). Details of these datasets are in the supplementary.


\subsection{Experimental settings}

\keypoint{Monitoring overfitting} We use parameters $N_m, N_w$ to say over how many iterations to calculate the moving average and how many iterations to wait before reset since the best moving average value. We select $N_m=N_w$ and use a value of 50 steps in most cases, while we use 100 for CIFAR-100 and Kuzushiji-49, and 200 for other larger-scale experiments (more than 100 base examples). These parameters do not affect the total number of iterations.

\keypoint{Early stopping for learning synthetic labels} We update the synthetic labels for a given number of epochs and then select the best labels to use based on the validation performance. For this, we train a new model from scratch using the current distilled labels and the associated base examples and then evaluate the model on the validation part of the real training set. We randomly set aside about 10-15\% (depending on the dataset) of the training data for validation. 

\keypoint{Models} We use LeNet \citep{LeCun1998Gradient-basedRecognition} for MNIST and similar experiments, and AlexNet \citep{Krizhevsky2012ImageNetNetworks} for CIFAR-10, CIFAR-100 and CUB. Both models are identical to the ones used in \citep{Wang2018DatasetDistillation}. In a fully supervised setting they achieve about 99\% and 80\% test accuracy on MNIST and CIFAR-10.

\keypoint{Selection of base examples} The base examples are selected randomly, using a shared random seed for consistency across scenarios. Our baseline models use the same random seed as the distillation models, so they share base examples for fair comparison. For within-dataset label distillation, we create a balanced set of base examples, so each class has the same number of base examples. For cross-dataset label distillation, we do not consider the original classes of base examples. The size of the label space and the labels are different in the source and the target problem. Our additional analysis (Tables \ref{tab:variousbaseexamples} and \ref{tab:optimizedbaseexamples} in the supplementary) has shown that the specific random set of base examples does not have a significant impact on the success of label distillation.

\begin{table*}[tb]
\caption{Within-dataset distillation recognition accuracy (\%). Our label distillation (LD) outperforms prior dataset distillation (DD) \cite{Wang2018DatasetDistillation} and soft-label dataset distillation (SLDD) \cite{Sucholutsky2019Soft-LabelDistillation}, while allowing significantly more flexible training of a new model as shown in Section \ref{flexsec}.}
\label{tab:ddproblem1}
\centering
\resizebox{1.0\textwidth}{!}{
\begin{tabular}{clcccccc}
\toprule
&Base examples & 10 & 20 & 50 & 100 & 200 & 500\\
\midrule
\parbox[t]{2mm}{\multirow{6}{*}{\rotatebox[origin=c]{90}{MNIST}}} &  LD & 60.89 $\pm$ 3.20 & 74.37 $\pm$ 1.27 & 82.26 $\pm$ 0.88 & \textbf{87.27 $\pm$ 0.69} & 91.47 $\pm$ 0.53 & 93.30 $\pm$ 0.31 \\
& Baseline & 48.35 $\pm$ 3.03 & 62.60 $\pm$ 3.33 & 75.07 $\pm$ 2.40 & 82.06 $\pm$ 1.75 & 85.95 $\pm$ 0.98 & 92.10 $\pm$ 0.43 \\
& Baseline LS & 51.22 $\pm$ 3.18 & 64.14 $\pm$ 2.57 & 77.94 $\pm$ 1.26 & 85.95 $\pm$ 1.09 & 90.10 $\pm$ 0.60 & 94.75 $\pm$ 0.29 \\
& LD RR & 64.57 $\pm$ 2.67 & 75.98 $\pm$ 1.00 & 82.49 $\pm$ 0.93 & \textbf{87.85 $\pm$ 0.43} & 88.88 $\pm$ 0.28 & 89.91 $\pm$ 0.33 \\
& Baseline RR & 52.53 $\pm$ 2.61 & 60.44 $\pm$ 1.97 & 74.85 $\pm$ 2.37 & 81.40 $\pm$ 2.11 & 87.03 $\pm$ 0.69 & 92.10 $\pm$ 0.80 \\
& Baseline RR LS & 51.53 $\pm$ 2.42 & 60.91 $\pm$ 1.78 & 76.26 $\pm$ 1.80 & 83.13 $\pm$ 1.41 & 87.94 $\pm$ 0.67 & 93.48 $\pm$ 0.61 \\
& DD \cite{Wang2018DatasetDistillation}  &  &  & & 79.5\phantom{0} $\pm$ 8.1\phantom{0} & & \\
& SLDD \cite{Sucholutsky2019Soft-LabelDistillation} &  &  & & 82.7\phantom{0} $\pm$ 2.8\phantom{0} & & \\
\midrule
\parbox[t]{2mm}{\multirow{6}{*}{\rotatebox[origin=c]{90}{CIFAR-10}}} &  LD & 25.69 $\pm$ 0.72 & 30.00 $\pm$ 0.86 & 35.36 $\pm$ 0.64 & \textbf{38.33 $\pm$ 0.44} & 41.05 $\pm$ 0.71 & 42.45 $\pm$ 0.40 \\
& Baseline & 14.29 $\pm$ 1.40 & 16.80 $\pm$ 0.72 & 20.75 $\pm$ 1.05 & 25.76 $\pm$ 1.04 & 31.53 $\pm$ 1.02 & 38.33 $\pm$ 0.75 \\
& Baseline LS & 13.22 $\pm$ 1.22 & 18.36 $\pm$ 0.65 & 22.81 $\pm$ 0.71 & 27.27 $\pm$ 0.68 & 33.62 $\pm$ 0.81 & 39.22 $\pm$ 1.12 \\
& LD RR & 25.07 $\pm$ 0.69 & 29.83 $\pm$ 0.46 & 35.23 $\pm$ 0.64 & \textbf{37.94 $\pm$ 1.22} & 41.17 $\pm$ 0.33 & 43.16 $\pm$ 0.47 \\
& Baseline RR & 13.37 $\pm$ 0.79 & 17.08 $\pm$ 0.31 & 19.85 $\pm$ 0.51 & 24.65 $\pm$ 0.47 & 28.97 $\pm$ 0.74 & 36.31 $\pm$ 0.49 \\
& Baseline RR LS & 13.82 $\pm$ 0.85 & 16.95 $\pm$ 0.52 & 20.00 $\pm$ 0.57 & 24.84 $\pm$ 0.60 & 29.28 $\pm$ 0.56 & 35.73 $\pm$ 1.02 \\
& DD \cite{Wang2018DatasetDistillation} &  &  &  & 36.8\phantom{0} $\pm$ 1.2\phantom{0} & & \\
& SLDD \cite{Sucholutsky2019Soft-LabelDistillation} &  &  &  & \textbf{39.8}\phantom{0} $\pm$ 0.8\phantom{0} & & \\
\bottomrule
\end{tabular}}
\end{table*}


\keypoint{Further details} Outer-loop minibatch uses $N_o=1024$ examples, while the inner minibatch size $N_i$ depends on the number of base examples. For 100 or more base examples, we use a minibatch of 50 examples, except for CIFAR-100 for which we use 100 examples. For 10, 20 and 50 base examples our minibatch sizes are 10, 10 and 25. We optimize the synthetic labels and the model using Adam optimizer with standard parameters ($\beta=0.001$). Most models are trained for 400 epochs, while larger-scale models (more than 100 base examples and CIFAR-100) are trained for 800 epochs. Smaller-scale Kuzushiji-49 experiments are trained for 100 epochs, while larger-scale ones use 200 epochs. Epochs are calculated based on the number of real training set examples, rather than base examples. In the second-order version, we do one inner-loop step update, using a learning rate of $\alpha=0.01$. We back-propagate through the inner-loop update when updating the synthetic labels (meta-knowledge), but not when subsequently updating the model $\theta$ with Adam optimizer. In the RR version, we use a pseudo-gradient step size $\alpha$ of 0.01 and regularization parameter $\lambda$ of 1.0. We calibrate the regression weights by scaling them with a value learned during training with the specific set of base examples and distilled labels. 
Our tables report the mean test accuracy and standard deviation (\%) across 20 models trained from scratch using the base examples and synthetic labels.

\begin{wraptable}{r}{0.3\textwidth}
\caption{CIFAR-100 within-dataset distillation. One example per class.}
\label{tab:cifar100}
\centering
\resizebox{0.3\textwidth}{!}{
\begin{tabular}{lcccc}
\toprule
LD & 11.46 $\pm$ 0.39 \\
Baseline & \phantom{0}3.51 $\pm$ 0.31 \\
Baseline LS & \phantom{0}4.07 $\pm$ 0.23 \\
LD RR & 10.80 $\pm$ 2.36 \\
Baseline RR & \phantom{0}3.00 $\pm$ 0.39 \\
Baseline RR LS & \phantom{0}3.65 $\pm$ 0.28 \\
\bottomrule
\end{tabular}}
\end{wraptable}
\subsection{Within-dataset distillation}
We compare our label distillation (LD) to previous dataset distillation (DD) and soft-label dataset distillation (SLDD) on MNIST and CIFAR-10. We also establish new baselines that take true labels from the target dataset and otherwise are trained in the same way as LD models. RR baselines use RR and pseudo-gradient for consistency with LD RR (overall network architecture remains the same as in the second-order approach). In addition, we include baselines that use label smoothing (LS) \cite{Szegedy2016RethinkingVision} with a smoothing parameter of 0.1 as suggested in \citep{Pereyra2017RegularizingDistributions}. The number of training steps for our baselines is optimized using the validation set, by training a model for various numbers of steps between 10 and 1000 and measuring the validation set accuracy (up to 1700 steps are used for cases with more than 100 base examples). Table \ref{tab:ddproblem1} shows that LD significantly outperforms previous work on MNIST. This is in part due to LD enabling the use of more steps (LD estimates $T_i\approx200-300$ steps vs fixed 3 epochs of 10 steps in LD and SLDD). Our improved baselines are also competitive, and outperform the prior baselines in \cite{Wang2018DatasetDistillation} due to taking more steps. The comparison is reasonable as it is very expensive to use DD and SLDD with many more steps. The standard uniform label smoothing baseline works well on MNIST for a large number of base examples, where the problem anyway approaches one of conventional supervised learning. However, this strategy has not shown to be effective enough for CIFAR-10, where synthetic labels are the best. Importantly, in the most intensively distilled regime of 10 examples, LD clearly outperforms all competitors. We provide an analysis of the labels learned by our method in Section \ref{sec: furtheranalysis}.
For CIFAR-10 our results also improve on the original DD result. In this experiment, our second-order algorithm performs similarly to our RR pseudo-gradient strategy.

We further show in Table \ref{tab:cifar100} that our distillation approach scales to a significantly larger number of classes than 10 by application to the CIFAR-100 benchmark. As before, we establish new baseline results that use the original labels (or their smoother alternatives) of the same images as those used as base examples in distillation. We use the validation set to choose a suitable number of steps for training a baseline, allowing up to 1000 steps, which is significantly more than what is typically used by LD. The results show our distillation method leads to clear improvements over the baseline.

\begin{table*}[t]
\caption{Cross-dataset distillation recognition accuracy (\%). Datasets: E = EMNIST, M = MNIST, K = KMNIST, B = CUB, C = CIFAR-10, K-49 = Kuzushiji-49.}
\label{tab:diffprob1}
\centering
\resizebox{1.0\textwidth}{!}{
\begin{tabular}{lcccccc}
\toprule
Base examples & 10 & 20 & 50 & 100 & 200 & 500 \\
\midrule
E $\rightarrow$ M (LD) & 36.13 $\pm$ 5.51 & 57.82 $\pm$ 1.91 & 69.21 $\pm$ 1.82 & 77.09 $\pm$ 1.66 & 83.67 $\pm$ 1.43 & 86.02 $\pm$ 1.00 \\
E $\rightarrow$ M (LD RR) & 56.08 $\pm$ 2.88 & 67.62 $\pm$ 3.03 & 80.80 $\pm$ 1.44 & 82.70 $\pm$ 1.33 & 84.44 $\pm$ 1.18 & 86.79 $\pm$ 0.76 \\
\midrule
E $\rightarrow$ K (LD) & 31.90 $\pm$ 2.82 & 39.88 $\pm$ 1.98 & 47.93 $\pm$ 1.38 & 51.91 $\pm$ 0.85 & 57.46 $\pm$ 1.37 & 59.84 $\pm$ 0.80 \\
E $\rightarrow$ K (LD RR) & 34.35 $\pm$ 3.37 & 46.67 $\pm$ 1.66 & 53.13 $\pm$ 1.88 & 57.02 $\pm$ 1.24 & 58.01 $\pm$ 1.28 & 63.77 $\pm$ 0.71 \\
\midrule
B $\rightarrow$ C (LD) & 26.86 $\pm$ 0.97 & 28.63 $\pm$ 0.86 & 31.21 $\pm$ 0.74 & 34.02 $\pm$ 0.66 & 38.39 $\pm$ 0.56 & 38.12 $\pm$ 0.41 \\
B $\rightarrow$ C (LD RR) & 26.50 $\pm$ 0.54 & 28.95 $\pm$ 0.47 & 32.23 $\pm$ 3.59 & 32.19 $\pm$ 7.89 & 36.55 $\pm$ 6.32 & 38.46 $\pm$ 6.97 \\
\midrule
E $\rightarrow$ K-49 (LD) & \phantom{0}7.37 $\pm$ 1.01 & \phantom{0}9.79 $\pm$ 1.23 & 17.80 $\pm$ 0.78 & 19.17 $\pm$ 1.27 & 22.78 $\pm$ 0.98 & 23.99 $\pm$ 0.81 \\
E $\rightarrow$ K-49 (LD RR) & 10.48 $\pm$ 1.26 & 14.84 $\pm$ 1.83 & 21.59 $\pm$ 1.87 & 20.86 $\pm$ 1.81 & 24.59 $\pm$ 2.26 & 24.72 $\pm$ 1.78 \\
\bottomrule
\end{tabular}}
\end{table*}



\subsection{Cross-dataset task}
For cross-dataset distillation, we considered four scenarios: from EMNIST letters to MNIST digits, from EMNIST letters (``English'') to Kuzushiji-MNIST or Kuzushiji-49 characters (``Japanese'') and from CUB bird species to CIFAR-10 general categories. Table \ref{tab:diffprob1} shows we are able to distill labels on examples of a different source dataset and achieve surprisingly good performance on the target problem, given no target data is used when training these models. In contrast, directly applying a trained source-task model to the target without distillation unsurprisingly leads to chance performance (about 2\% test accuracy for Kuzushiji-49 and 10\% for all other cases). These results show we can indeed distill the knowledge of one dataset into base examples from a different but related dataset through crafting synthetic labels. Furthermore, our RR approach surpasses the second-order method in most cases, confirming its value. When using 10 and 20 base examples for Kuzushiji-49, the number of training examples is smaller than the number of classes (49), providing a novel example of \emph{less-than-one-shot learning} where there are fewer examples than classes \cite{Sucholutsky2020LessSamples}.


\subsection{Flexibility of distilled datasets}
\label{flexsec}
We verify the flexibility of our label-distilled dataset compared to image-distilled alternative by \citet{Wang2018DatasetDistillation}. We look at: (1) How the number of steps used during meta-testing affects the accuracy of learning with distilled data, and in particular sensitivity to deviation from the number of steps used during meta-training of DD and LD. (2) Sensitivity of the models to changes in optimization parameters between meta-training and meta-testing. (3) How well the distilled datasets transfer to architectures different to those used for training. We used the DD implementation of \citet{Wang2018DatasetDistillation} for fair evaluation. Figure~\ref{fig:flex} summarizes the key points, and detailed tables are in the supplementary.

\begin{figure}[h!]
\centering
\begin{subfigure}{0.29\textwidth}
    \centering
    \includegraphics[width = \textwidth]{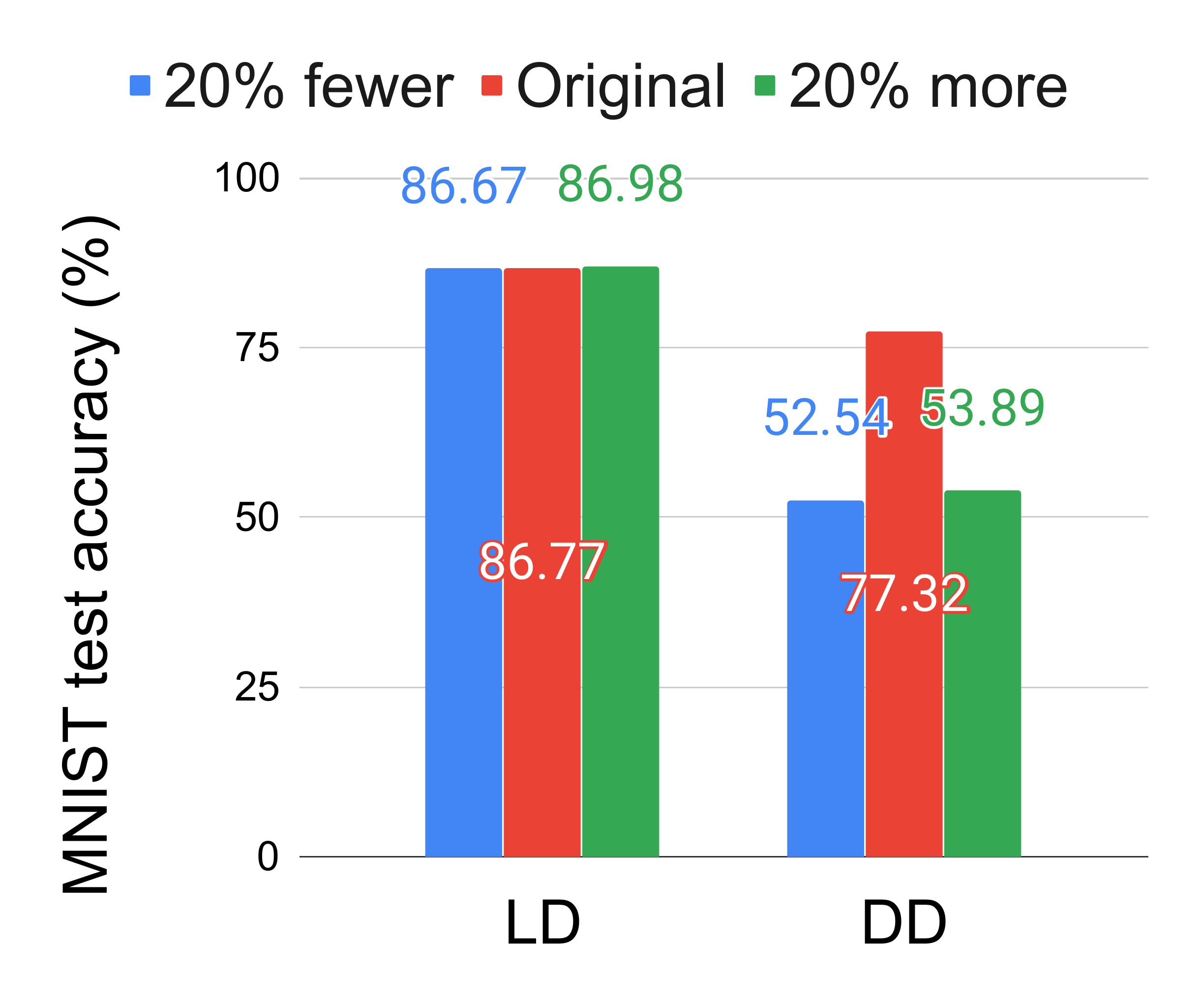}
    \caption{Optimization steps}
\end{subfigure}
\begin{subfigure}{0.29\textwidth}
    \centering
    \includegraphics[width = \textwidth]{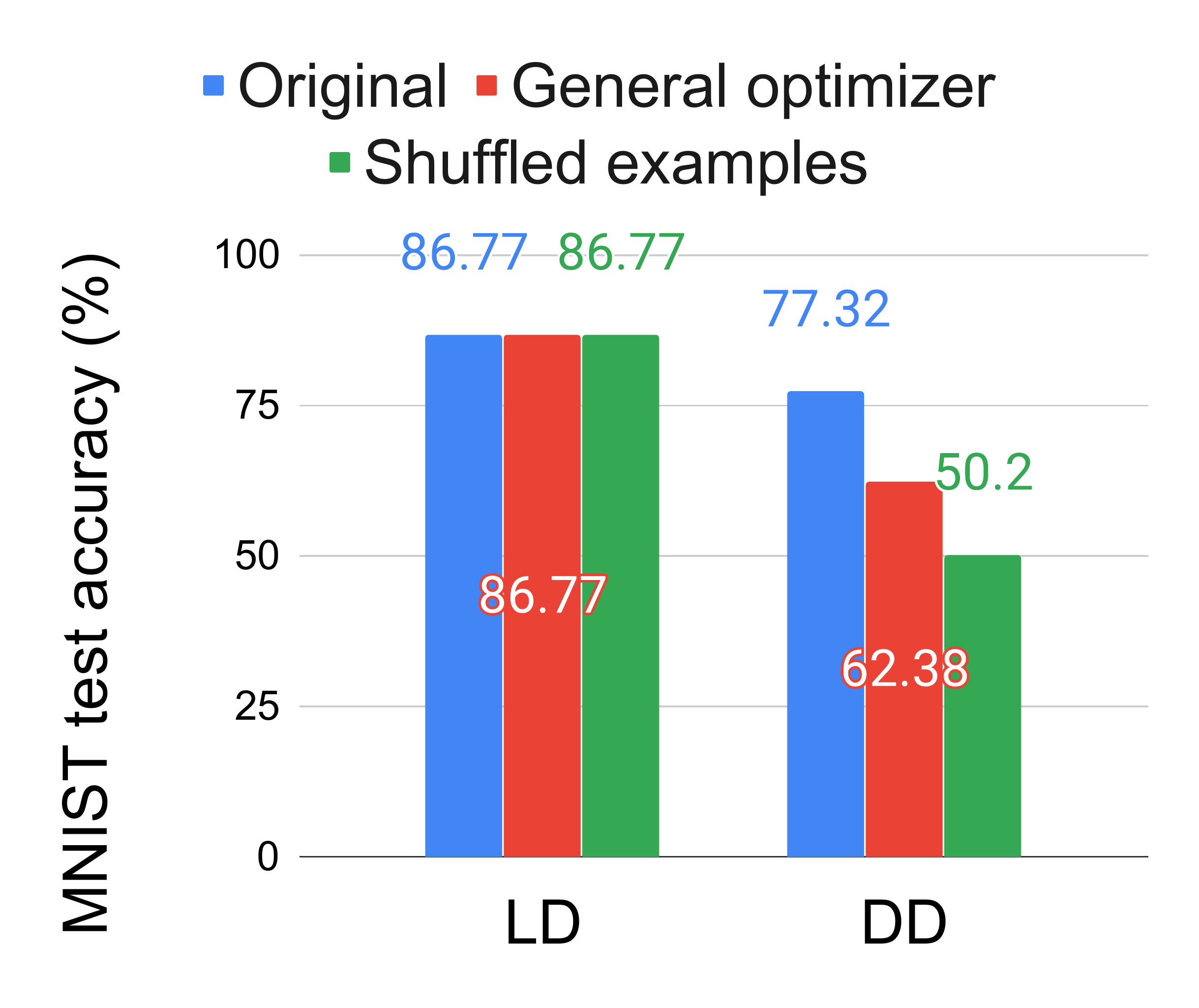}
    \caption{Optimization parameters}
\end{subfigure}
\begin{subfigure}{0.29\textwidth}
    \centering
    \includegraphics[width=\textwidth]{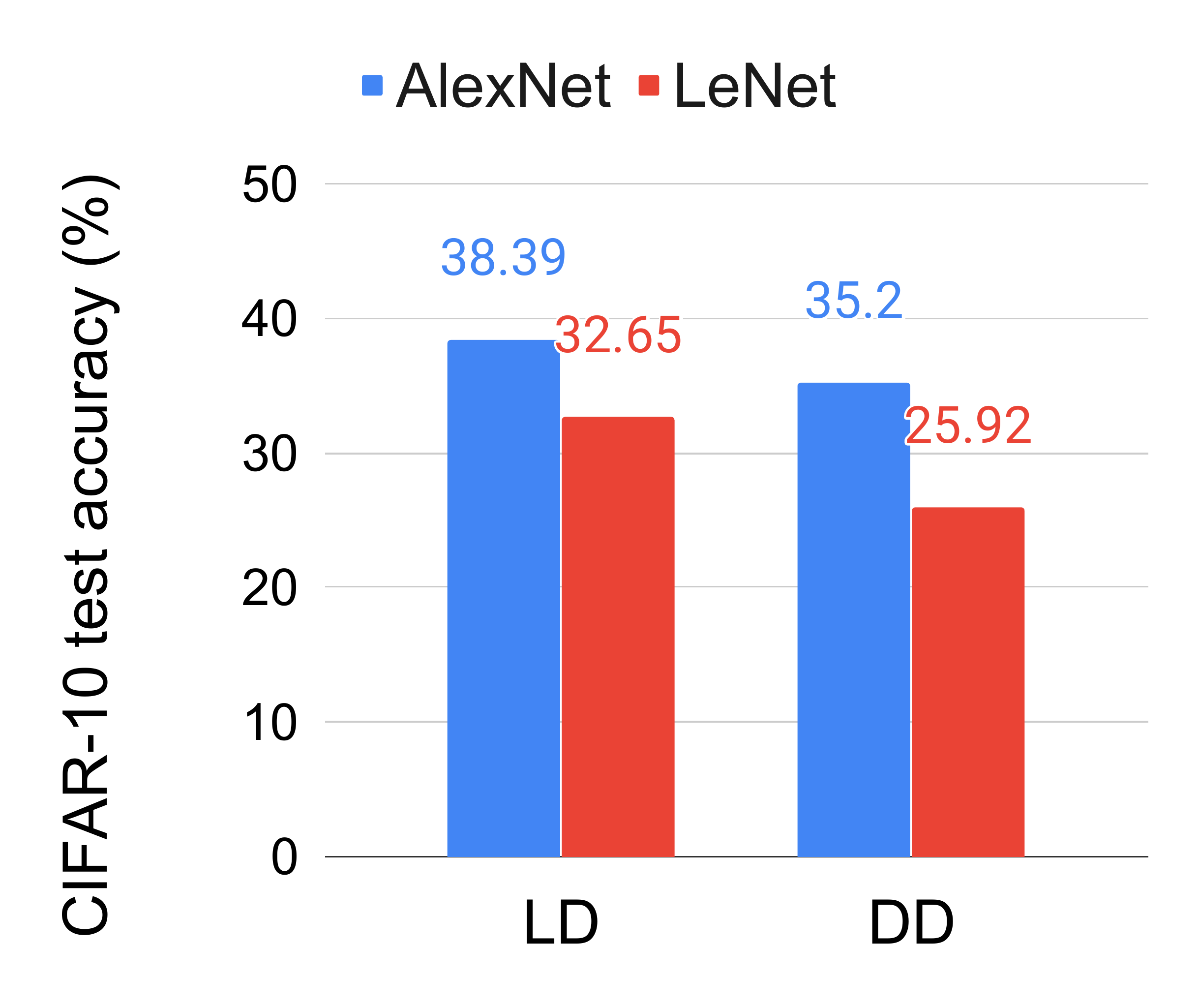}
    \caption{Cross-architecture transfer}
\end{subfigure}
\caption{Datasets distilled with LD are more flexible than those distilled with DD. \label{fig:flex}}
\end{figure}

\keypoint{Sensitivity to the number of meta-testing steps} (Figure~\ref{fig:flex}a)
Our method is relatively insensitive to the number of steps used to train a model on the distilled data. Table \ref{tab:numstepsanalysis} shows that even if we do 50 steps less or 100 steps more than the number estimated during training ($\approx 300$), test accuracy does not change significantly. However, previous DD and SLDD methods need to be trained for a specific number of steps with optimized learning rates. If the number of steps changes even by as little as 20\%, they incur a significant cost in accuracy. Table \ref{tab:origddstepssensitivity} provides a further sensitivity analysis of DD.

\keypoint{Sensitivity to optimization parameters} (Figure~\ref{fig:flex}b) DD uses step-wise meta-optimized learning rates to maximize accuracy. Table \ref{tab:origddlrsensitivity} shows using the average of the optimized learning rate rather than the specific value in each step (more general optimizer) leads to a significantly worse result. Original DD also relies on training the distilled data in a fixed sequence, and Table \ref{tab:origddordersensitivity} shows changing the order of examples leads to a large decrease in accuracy. Our LD method by design does not depend on the specific order of examples and can be used with off-the-shelf optimizers such as Adam.

\keypoint{Transferability of labels across architectures}(Figure~\ref{fig:flex}c)
We study the impact of distilling the labels using AlexNet and then training AlexNet, LeNet and ResNet-18 using the distilled labels. Tables \ref{tab:archor2} and \ref{tab:archrr} suggest our labels are transferable in both within and across dataset distillation scenarios. Table \ref{tab:diffarchdd} further shows original DD images are somewhat transferable if using the specific order of examples and optimized learning rates (the reported baseline results are worse). However, the decrease in test accuracy is smaller for our LD, suggesting LD has better transferability.


\subsection{Further analysis}
\label{sec: furtheranalysis}

\keypoint{Analysis of synthetic labels} We have analysed to what extent the synthetic labels learn the true label (Figures \ref{fig:labelsanalysisc10} and \ref{fig:labelsanalysis} in the supplementary). Our RR method has typically led to more complex labels that recovered the true label to a smaller extent than the second-order version (e.g. 63\% vs 84\% on the same scenario). For cross-dataset LD, Figure \ref{fig:reconstructed_images} suggests LD can be understood as learning labels such that the base examples' label-weighted combination resembles the images from the target class. Example synthetic labels are included in the supplementary.

\begin{wrapfigure}{r}{0.34\textwidth}
    \centering
    \includegraphics[width =0.34 \textwidth]{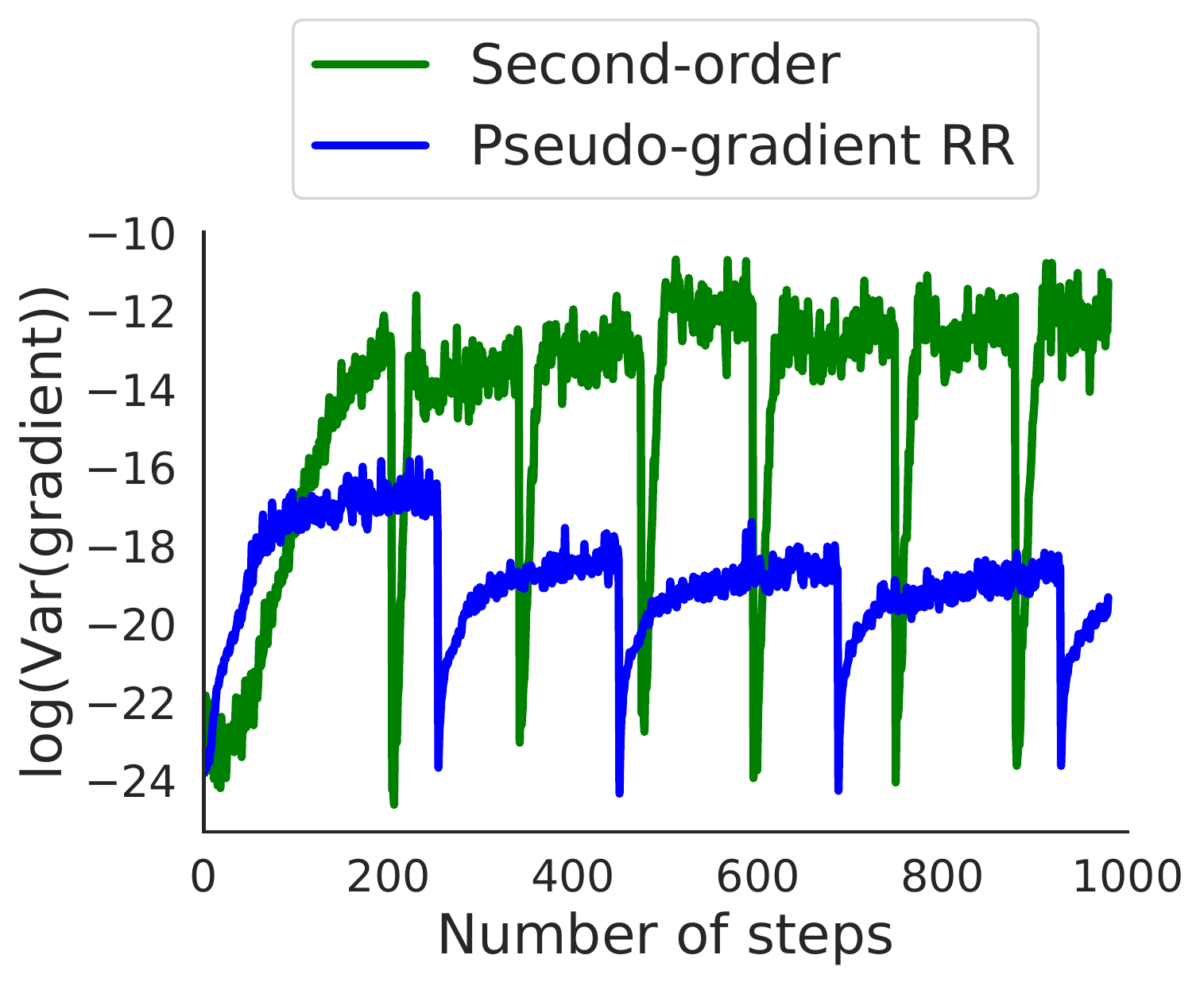}
    \caption{RR-based LD reduces synthetic label meta-gradient variance $\mathrm{Var} \left[ \nabla_{\tilde{\boldsymbol{Y}}_\mathcal{S}} L \right]$.}
\label{fig:gradvar}
\end{wrapfigure}


\keypoint{Pseudo-gradient analysis} Our pseudo-gradient RR method obtains a significantly lower variance of meta-knowledge gradients than the second-order method, as shown in Figure \ref{fig:gradvar}. This leads to more stable and effective training.

\keypoint{Discussion} LD provides a more effective and more flexible distillation approach than prior work. This brings us one step closer to the vision of leveraging distilled data to accelerate model training, or design -- such as architecture search \citep{Elsken2019NeuralSurvey}. Currently we only explicitly randomize over network initializations during training. In future work we believe our strategy of multi-step training and reset at convergence could be used with other factors, such as randomly selected network architectures to further improve cross-network generalisation performance.

\section{Conclusion}
We have introduced a new label distillation algorithm for distilling the knowledge of a large dataset into synthetic labels of a few base examples from the same or a different dataset. Our method improves on prior dataset distillation results, scales well to larger problems, and enables novel settings such as cross-dataset distillation. Most importantly, it is significantly more flexible in terms of distilling general purpose datasets that can be used downstream with off-the-shelf optimizers.

\section*{Broader Impact}
We propose a flexible and efficient distillation scheme that gets us closer to the goal of practically useful dataset distillation. Dataset distillation could ultimately lead to a beneficial impact in terms of researcher's time efficiency by enabling faster experimentation when training on small distilled datasets. Perhaps more importantly, it could reduce the environmental impact of AI research and development by reduction of energy costs \citep{Schwartz2019GreenAI}. However, our results are still not strong enough yet. For this goal to be realised better distillation methods leading to more accurate downstream models need to be developed. 

In the future, label distillation could speculatively provide a useful tool for privacy preserving learning \citep{Al-Rubaie2019Privacy-PreservingSolutions}, for example in situations where organizations want to learn from user's data. A company could provide a small set of public (source) data to a user, who performs cross-dataset distillation using their private (target) data to train a model. The user would then return the distilled labels on the public data, which would allow the company to re-create the user's model. In this way the knowledge from the user's training could be obtained by the company in the form of distilled labels -- without directly sending any private data, or trained model that could be a vector for memorization attacks \citep{Carlini2019TheNetworks}.

On the negative side, detecting and understanding the impact of bias in datasets is an important yet already very challenging issue for machine learning. The impact of dataset distillation on any underlying biases in the data is completely unclear. If people were to train models on distilled datasets in the future, it would be important to understand the impact of distillation on data biases.

\section*{Source Code}
We provide a PyTorch implementation of our approach at \url{https://github.com/ondrejbohdal/label-distillation}.

\begin{ack}
This work was supported in part by the EPSRC Centre for Doctoral Training in Data Science, funded by the UK Engineering and Physical Sciences Research Council (grant EP/L016427/1) and the University of Edinburgh.


\end{ack}

\bibliographystyle{apalike}
\bibliography{references,extra}

\clearpage
\appendix

\section{Datasets}
\label{sec:datasets}
\begin{figure}[h!]
\begin{center}
\centerline{\includegraphics[width=0.8\columnwidth]{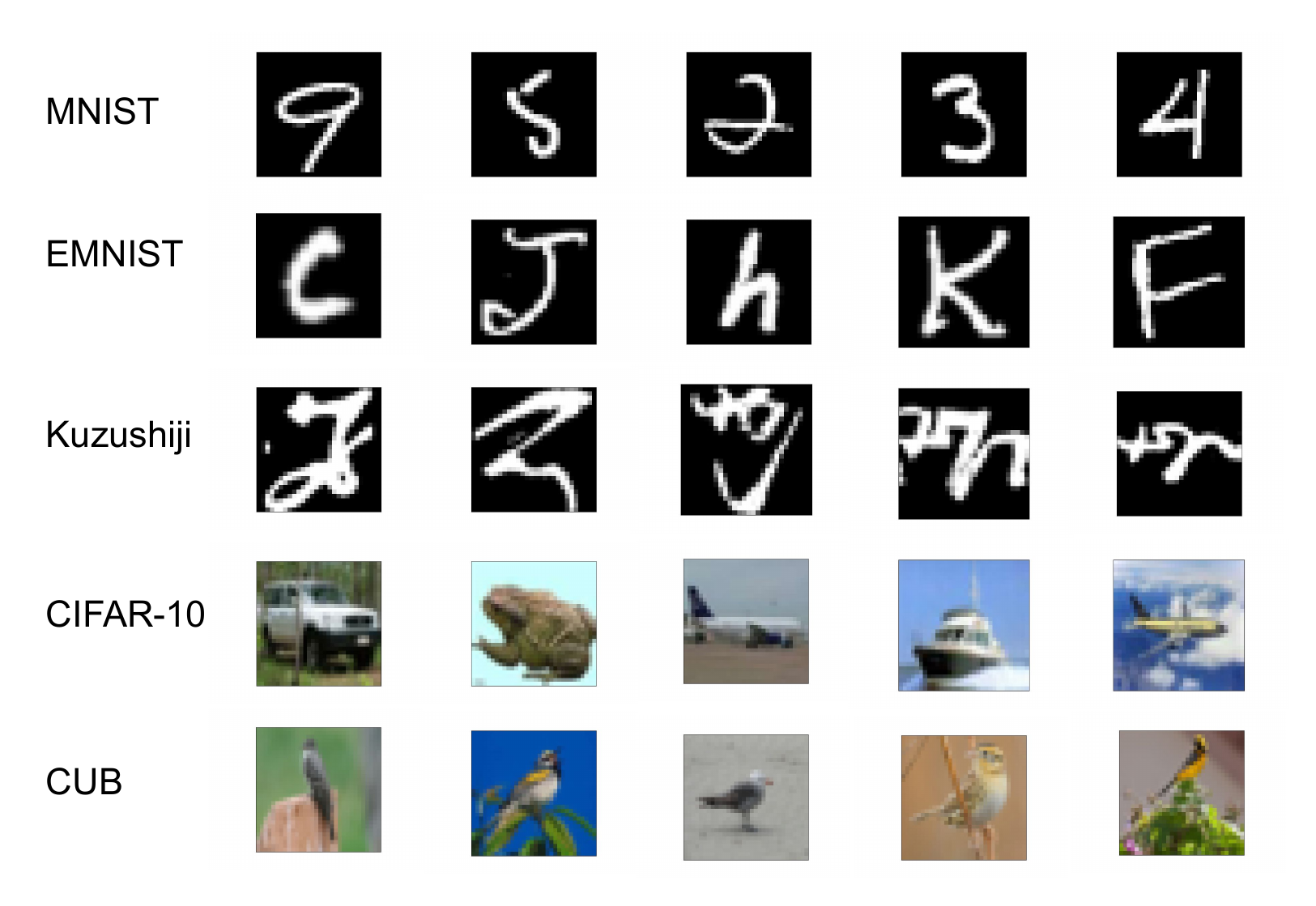}}
\caption{Example images from the different datasets that we use.}
\label{fig:DatasetsIllustration}
\end{center}
\end{figure}

We use MNIST \citep{LeCun1998Gradient-basedRecognition}, EMNIST \citep{Cohen2017EMNIST:Letters}, KMNIST and Kuzushiji-49 \citep{Clanuwat2018DeepLiterature}, CIFAR-10 and CIFAR-100 \citep{Krizhevsky2009LearningImages}, and CUB \citep{Wah2011TheDataset} datasets. Example images are shown in Figure~\ref{fig:DatasetsIllustration}. MNIST includes images of 70000 handwritten digits that belong into 10 classes. EMNIST dataset includes various characters, but we choose EMNIST letters split that includes only letters. Lowercase and uppercase letters are combined together into 26 balanced classes (145600 examples in total). KMNIST (Kuzushiji-MNIST) is a dataset that includes images of 10 classes of cursive Japanese (\textit{Kuzushiji}) characters and is of the same size as MNIST. Kuzushiji-49 is a larger version of KMNIST with 270912 examples and 49 classes. CIFAR-10 includes 60000 colour images of various general objects, for example airplanes, frogs or ships. As the name indicates, there are 10 classes. CIFAR-100 is like CIFAR-100, but has 100 classes with 600 images for each of them. Every class belongs to one of 20 superclasses which represent more general concepts. CUB includes colour images of 200 bird species. The number of images is small, only 11788. All datasets except Kuzushiji-49 are balanced or almost balanced.

\section{Analysis of simple one-layer case}
In this section we analyse how synthetic labels are meta-learned in the case of a simple one-layer model with sigmoid output layer $\sigma$, second-order approach and binary classification problem. We will consider one example at a time for simplicity. The model has weights $\boldsymbol{\theta}$ and gives prediction $\hat{y}=\sigma(\boldsymbol{\theta}^T\boldsymbol{x})$ for input image $\boldsymbol{x}$ with true label $y$. We use binary cross-entropy loss $L$: $$L\left(\hat{y},y\right)=-y\log\hat{y}-\left(1-y\right)\log\left(1-\hat{y}\right).$$ 

As part of the algorithm, we first update the base model, using the current base example and synthetic label:
$$\boldsymbol{\theta}^\prime = \boldsymbol{\theta} - \alpha\nabla_{\boldsymbol{\theta}} L\left(\sigma(\boldsymbol{\theta}^T\boldsymbol{\Tilde{x}}), \Tilde{y}\right),$$ after which we update the synthetic label: 
$$\Tilde{y} \leftarrow \Tilde{y} - \beta\nabla_{\Tilde{y}} L\left(\sigma(\boldsymbol{\theta}^{\prime T}\boldsymbol{x}), y\right).$$ Notation: $\boldsymbol{\Tilde{x}}$ is the base example, $\Tilde{y}$ is the synthetic label, $\alpha$ is the inner-loop learning rate, $\beta$ is the outer-loop learning rate, $\boldsymbol{x}$ is an example from the target set, $y$ is the label of the example and $\boldsymbol{\theta}$ describes the model weights.

Our goal is to intuitively interpret the update of the synthetic label, which uses the gradient $\nabla_{\Tilde{y}} L\left(\sigma(\boldsymbol{\theta}^{\prime T}\boldsymbol{x}), y\right)$. We will repeatedly use the chain rule and the fact that $$\frac{\partial \sigma (x)}{\partial x}=\sigma (x) \left(1-\sigma (x)\right).$$ Moreover, we will use the following result (for binary cross-entropy loss $L$ introduced earlier):
\begin{equation*} \label{lossgrad}
\begin{split}
\frac{\partial L\left(\sigma(\boldsymbol{\theta}^T\boldsymbol{x}), y\right)}{\partial \boldsymbol{\theta}}&=\frac{\partial L\left(\sigma(\boldsymbol{\theta}^T\boldsymbol{x}), y\right)}{\partial \sigma(\boldsymbol{\theta}^T\boldsymbol{x})} \frac{\partial \sigma(\boldsymbol{\theta}^T\boldsymbol{x})}{\partial \boldsymbol{\theta}} \\
&=\left(\sigma(\boldsymbol{\theta}^T\boldsymbol{x})-y\right)\boldsymbol{x}
\end{split}
\end{equation*}

Now we derive an intuitive formula for the gradient used for updating the synthetic label:
\begin{equation*} \label{eqgrad}
\begin{split}
\frac{\partial L\left(\sigma(\boldsymbol{\theta}^{\prime T}\boldsymbol{x}), y\right)}{\partial \tilde{y}}&=\frac{\partial L\left(\hat{y}^\prime, y\right)}{\partial \tilde{y}}=\left(\frac{\partial L\left(\hat{y}^\prime, y\right)}{\partial \hat{y}^\prime} \frac{\partial \hat{y}^\prime}{\partial \boldsymbol{\theta}^\prime}\right)^T \frac{\partial \boldsymbol{\theta}^\prime}{\partial \tilde{y}}\\
&= \left(\frac{\partial L\left(\hat{y}^\prime, y\right)}{\partial \hat{y}^\prime} \frac{\partial \hat{y}^\prime}{\partial \boldsymbol{\theta}^\prime}\right)^T \frac{\partial \left(\boldsymbol{\theta} - \alpha\nabla_{\boldsymbol{\theta}} L\left(\sigma(\boldsymbol{\theta}^T\boldsymbol{\Tilde{x}}), \Tilde{y}\right) \right)}{\partial \tilde{y}} \\
&=\left(\frac{\partial L\left(\hat{y}^\prime, y\right)}{\partial \hat{y}^\prime} \frac{\partial \hat{y}^\prime}{\partial \boldsymbol{\theta}^\prime}\right)^T \frac{\partial \left(\boldsymbol{\theta} - \alpha \left(\sigma(\boldsymbol{\theta}^T\boldsymbol{\Tilde{x}})-\tilde{y}\right)\boldsymbol{\Tilde{x}}\right)}{\partial \tilde{y}} \\
&=\left(\frac{\partial L\left(\hat{y}^\prime, y\right)}{\partial \hat{y}^\prime} \frac{\partial \hat{y}^\prime}{\partial \boldsymbol{\theta}^\prime}\right)^T \left(\alpha \boldsymbol{\Tilde{x}}\right) = \left(\left(\hat{y}^\prime-y\right)\boldsymbol{x}\right)^T \left(\alpha \boldsymbol{\Tilde{x}}\right) \\
&= \alpha \left( \sigma(\boldsymbol{\theta}^{\prime T}\boldsymbol{x}) - y\right)\boldsymbol{x}^T \boldsymbol{\Tilde{x}}
\end{split}
\end{equation*}

The next step is to interpret the update rule. The update is proportional to the difference between the prediction on the real training set and the true label $\left( \sigma(\boldsymbol{\theta}^{\prime T}\boldsymbol{x}) - y\right)$ as well as to the similarity between the real training set example and the base example $\left(\boldsymbol{x}^T \boldsymbol{\Tilde{x}}\right)$. This suggests the synthetic labels are updated so that they capture the different amount of similarity of a base example to examples from different classes in the target dataset. A similar analysis can also be done for our RR method -- in such case the result would be similar and would include a further proportionality constant dependent on the base examples (not affecting the intuitive interpretation).

\section{Additional experimental details}
\paragraph{Normalization} We normalize greyscale images using the standardly used normalization for MNIST (mean of 0.1307 and standard deviation of 0.3081). All our greyscale images are of size $28 \times 28$. Colour images are normalized using CIFAR-10 normalization (means of about 0.4914, 0.4822, 0.4465, and standard deviations of about 0.247, 0.243, 0.261 across channels). All colour images are reshaped to be of size $32 \times 32$.

\begin{wraptable}{r}{0.34\textwidth}
\caption{Comparison of training times of DD \cite{Wang2018DatasetDistillation} and our LD (mins).}
\label{tab:timecomparison}
\centering
\resizebox{0.34\columnwidth}{!}{
\begin{tabular}{lccc}
\toprule
 & MNIST & CIFAR-10 \\
\midrule
DD & 116 & 205 \\
LD & 61 & 86 \\
LD (RR) & 65 & 98 \\
Our DD & 67 & 96 \\
Our DD (RR) & 72 & 90 \\
\bottomrule
\end{tabular}}
\end{wraptable}

\paragraph{Computational resources} Each experiment was done on a single GPU, in almost all cases NVIDIA 2080 Ti. Shorter (400 epochs) experiments took about 1 or 2 hours to run, while longer (800 epochs) experiments took between 2 and 4 hours.

\paragraph{Training time} In Table~\ref{tab:timecomparison} we compare training times of our framework and the original DD (using the same settings and hardware). Besides evaluating LD, we also use our meta-learning algorithm to implement an image-distillation strategy for direct comparison with the original DD. The results show our online approach significantly accelerates training. Our DD is comparable to LD, and both are faster than original DD. However, we focus on LD because our version of DD was relatively unstable (Table \ref{tab:ddimagesanalysis}) and led to  worse performance than LD, perhaps because learning synthetic images is more complex than synthetic labels. This shows we need both the labels and our re-initializing strategy.

\begin{figure}[h!]
\begin{center}
\centerline{\includegraphics[width=0.65\columnwidth]{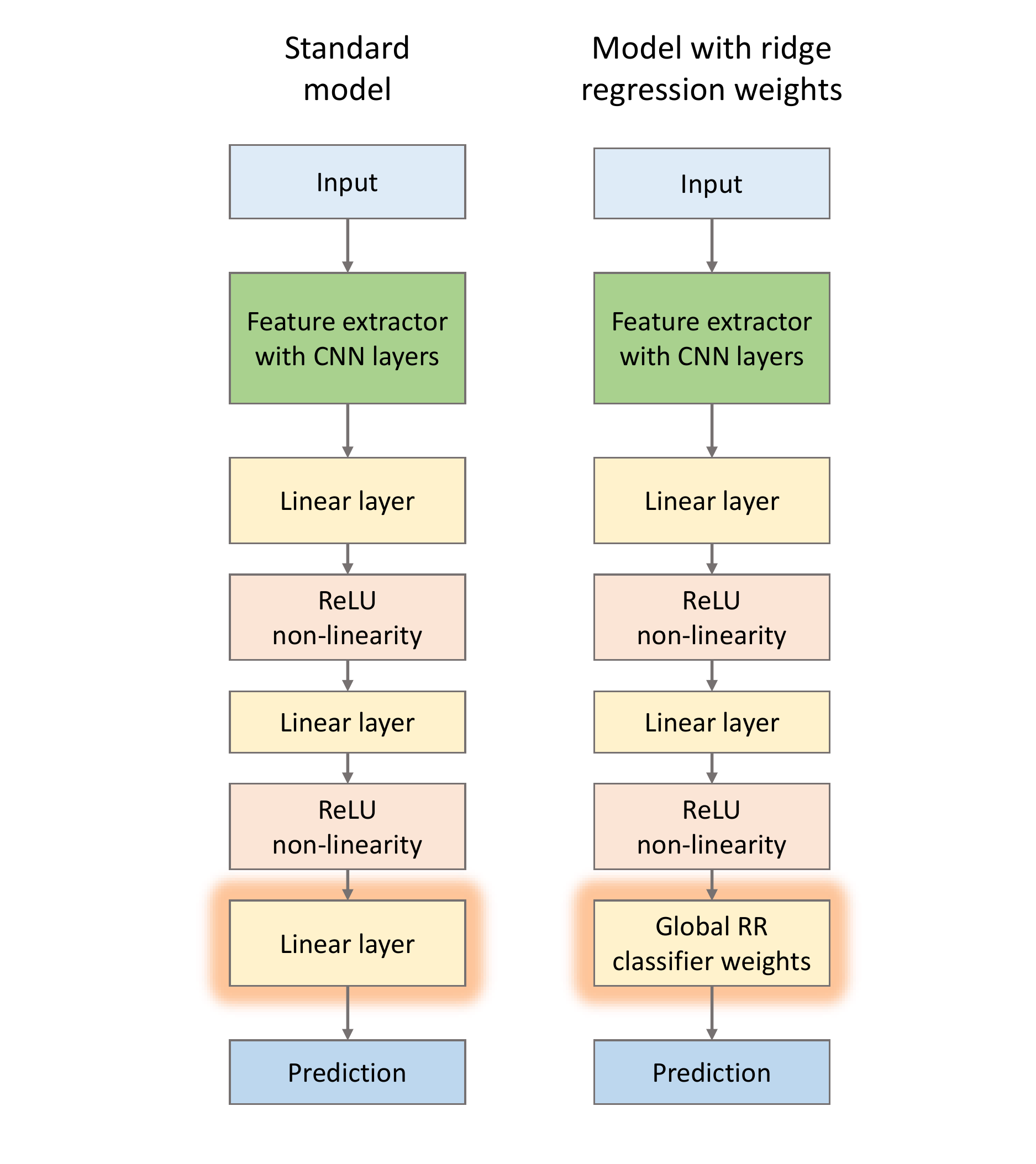}}
\vspace{-0.3cm}
\caption{Comparison of a standard model used for second-order label distillation and a model that uses global ridge regression classifier weights (used for first-order RR label distillation).}
\label{fig:StandardModelVsRRModelIllustration}
\end{center}
\vskip -5mm
\end{figure}

In addition, Figure \ref{fig:StandardModelVsRRModelIllustration} illustrates the difference between a standard model used for second-order label distillation and a model that uses global ridge regression classifier weights (used for first-order RR label distillation). The two models are almost identical -- only the final linear layer is different.

\section{Additional experiments}
\paragraph{Stability and dependence on choice of base examples} To evaluate the consistency of our results, we repeat the entire pipeline and report the results in Table~\ref{tab:trainallagain}. In the previous experiments, we used one randomly chosen but fixed set of base examples per source task. We investigate the impact of base example choice by drawing further random base example sets. The results in Table \ref{tab:variousbaseexamples} suggest that the impact of base example choice is slightly larger than that of the variability due to the distillation process, but still small overall. Note that the $\pm$ standard deviations in all cases quantify the impact of retraining from different random initializations at meta-test, given a fixed base set and completed distillation. It is likely that if the base examples were not selected randomly, the impact of using specific base examples would be larger. In fact, future work could investigate how to choose base examples so that learning synthetic labels for them improves the result further. We have tried the following strategy, but the label distillation results remained similar to the previous results:
\begin{itemize}
    \item Try 50 randomly selected sets of examples, train a model with each three times (for robustness) and measure the validation accuracy.
    \item The validation accuracy is measured for various numbers of steps, in most cases we evaluate every 50 steps up to 1000 steps (or 1700 steps when there are more than 100 base examples).
    \item Select the set with the largest mean validation accuracy at any point of training (across the three runs).
    \item This strategy maximizes the performance of the baselines, but could potentially also help the label distillation since these examples could be generally better for training.
\end{itemize}

The results for this strategy are in Table \ref{tab:optimizedbaseexamples}.

\paragraph{Dependence on target dataset size}
Our experiments use a relatively large target dataset (about 50000 examples) for meta-learning.
We study the impact of reducing the amount of target data for distillation in Table \ref{tab:targetsetsizevariable}. Using 5000 or more examples (about 10\% of the original size) is enough to achieve comparable performance.

\paragraph{Transferability of RR synthetic labels to standard model training}
When using RR, we train a validation and test model with RR and global classifier weights obtained using pseudo-gradient. In this experiment we study what happens if we create synthetic labels with RR, but do validation and testing with standard models trained from scratch without RR. For a fair comparison, we use the same synthetic labels for training a new RR model and a new standard model. Validation for early stopping is done with a standardly trained model. The results in Table \ref{tab:rrtransferability} suggest RR labels are largely transferable (even in cross-dataset scenarios), but there is some decrease in performance. Consequently, it is better to learn the synthetic labels using second-order approach if we want to train a standard model without RR during testing (comparing with the results in Table \ref{tab:ddproblem1}, \ref{tab:cifar100} and \ref{tab:diffprob1}).

\begin{figure*}[h!]
\centering
\begin{subfigure}{0.37\textwidth}
    \centering
    \includegraphics[width = \textwidth]{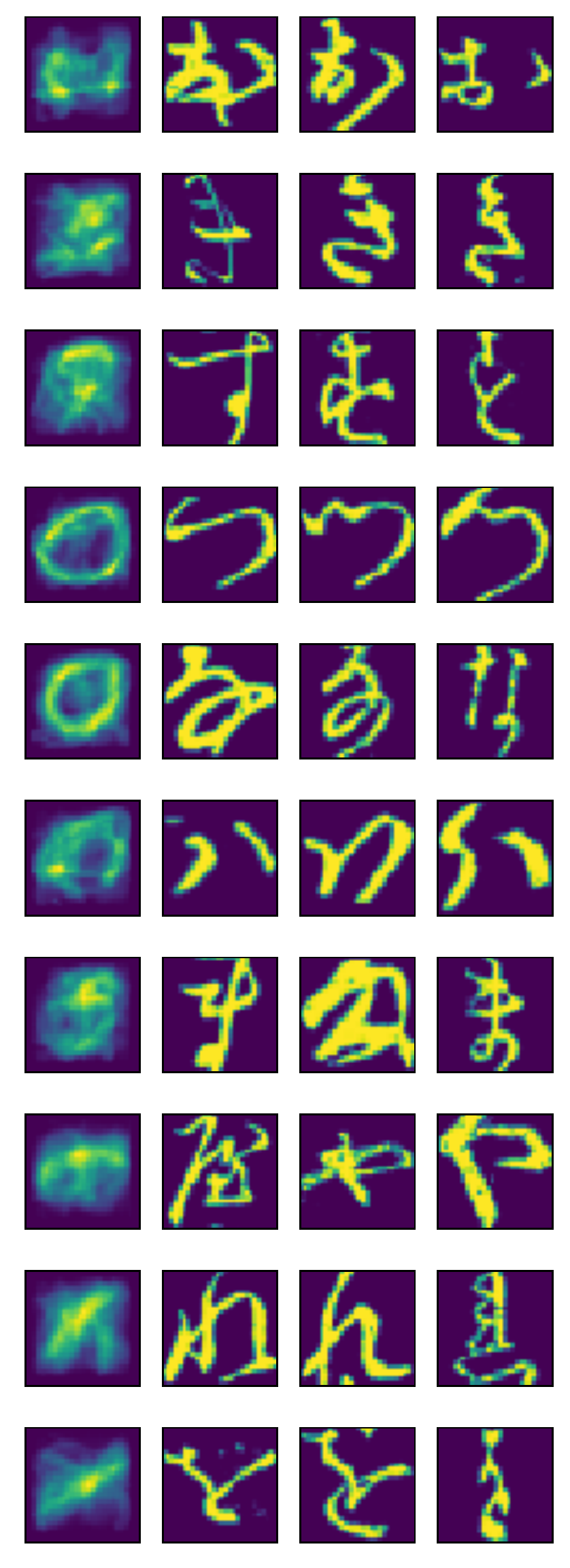}
    \caption{KMNIST}
\end{subfigure}
\begin{subfigure}{0.37\textwidth}
    \centering
    \includegraphics[width = \textwidth]{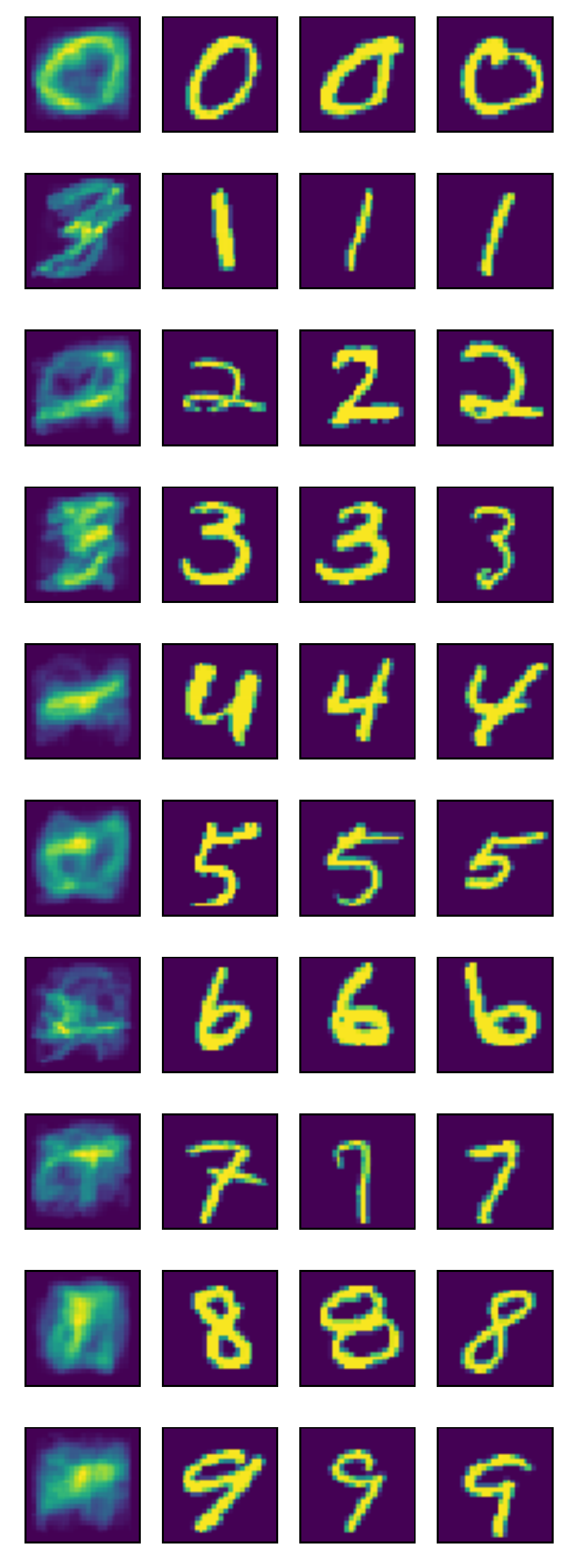}
    \caption{MNIST}
\end{subfigure}
\caption{Cross-dataset task: reconstructed images from KMNIST (Japanese letters) and MNIST (digits) based on combination of EMNIST base examples (English letters) (100 base examples used). Each row corresponds to a separate class, while the leftmost column shows the reconstructed image and the other three columns show actual examples from the same class. One image is reconstructed for each target dataset class. The base example images are combined pixel-wise with proportions based on the element of the synthetic label vector corresponding to the class that is being reconstructed. Note that KMNIST images in the same class can look very different because of different ways of writing the character, which makes reconstruction more challenging. Some of the reconstructed images resemble images from the target dataset classes, which shows that LD learns labels that combine base examples so that their pixel-wise combination, weighted based on the synthetic class labels, looks similar to the target class images. \label{fig:reconstructed_images}}
\end{figure*}

\paragraph{Intuition on cross-dataset distillation.} To illustrate the mechanism behind cross-dataset distillation, we use the distilled labels to linearly combine base EMNIST example images weighted by their learned synthetic labels in order to estimate a prototypical KMNIST/MNIST target class example as implied by learned LD labels. Although the actual mechanism is more complex than this due to the non-linearity of the neural network, we can qualitatively see individual  KMNIST/MNIST target classes are approximately encoded by their linear EMNIST LD prototypes as shown in Figure~\ref{fig:reconstructed_images}.

\section{Results of analysis}
Our tables report the mean test accuracy and standard deviation (\%) across 20 models trained from scratch using the base examples and synthetic labels. When analysing the original DD, 200 randomly initialized models are used.

\begin{table*}[h!]
\caption{Repeatability. Label distillation is quite repeatable. Performance change from repeating the whole distillation learning and subsequent re-training is small. We used 100 base examples for these experiments. Datasets: E = EMNIST, M = MNIST.}
\label{tab:trainallagain}
\centering
\begin{tabular}{lccc}
\toprule
 & Trial 1 & Trial 2 & Trial 3\\
\midrule
MNIST (LD) & 87.27 $\pm$ 0.69 & 87.49 $\pm$ 0.44 & 86.77 $\pm$ 0.77 \\
MNIST (LD RR) & 87.85 $\pm$ 0.43 & 88.31 $\pm$ 0.44 & 88.07 $\pm$ 0.46 \\
E $\rightarrow$ M (LD) & 77.09 $\pm$ 1.66 & 76.81 $\pm$ 1.47 & 77.10 $\pm$ 1.74 \\
E $\rightarrow$ M (LD RR) & 82.70 $\pm$ 1.33 & 83.06 $\pm$ 1.43 & 81.46 $\pm$ 1.70 \\
\bottomrule
\end{tabular}
\end{table*}

\begin{table*}[h!]
\caption{Base example sensitivity. Label distillation has some sensitivity to the specific set of base examples (chosen by a specific random seed), but the sensitivity is relatively low. We used 100 base examples for these experiments. It is likely that label distillation would be more sensitive for a smaller number of base examples.}
\label{tab:variousbaseexamples}
\centering
\resizebox{1.0\textwidth}{!}{
\begin{tabular}{lccccc}
\toprule
 & Set 1 & Set 2 & Set 3 & Set 4 & Set 5 \\
\midrule
MNIST (LD) & 84.91 $\pm$ 0.92 & 87.38 $\pm$ 0.81 & 87.49 $\pm$ 0.44 & 87.12 $\pm$ 0.47 & 85.16 $\pm$ 0.48 \\
MNIST (LD RR) & 87.82 $\pm$ 0.60 & 88.78 $\pm$ 0.57 & 88.31 $\pm$ 0.44 & 88.40 $\pm$ 0.46 & 87.77 $\pm$ 0.60 \\
E $\rightarrow$ M (LD) & 79.34 $\pm$ 1.36 & 74.55 $\pm$ 1.00 & 76.81 $\pm$ 1.47 & 78.59 $\pm$ 1.05 & 78.55 $\pm$ 1.32 \\
E $\rightarrow$ M (LD RR) & 81.67 $\pm$ 1.39 & 83.30 $\pm$ 1.38 & 83.06 $\pm$ 1.43 & 82.62 $\pm$ 1.70 & 83.43 $\pm$ 0.98 \\
\bottomrule
\end{tabular}}
\end{table*}

\begin{table*}[h!]
\caption{Optimized base examples: within-dataset distillation recognition accuracy (\%). Our label distillation (LD) outperforms prior Dataset Distillation \cite{Wang2018DatasetDistillation} (DD) and SLDD \cite{Sucholutsky2019Soft-LabelDistillation}, and scales to synthesizing more examples. The LD results remained similar to the original results even with optimized base examples.}
\label{tab:optimizedbaseexamples}
\centering
\resizebox{1.0\textwidth}{!}{
\begin{tabular}{clcccccc}
\toprule
&Base examples & 10 & 20 & 50 & 100 & 200 & 500\\
\midrule
\parbox[t]{2mm}{\multirow{6}{*}{\rotatebox[origin=c]{90}{MNIST}}} &  LD & 66.96 $\pm$ 2.01 & 74.37 $\pm$ 1.65 & 83.17 $\pm$ 1.28 & 86.66 $\pm$ 0.44 & 90.75 $\pm$ 0.49 & 93.22 $\pm$ 0.41 \\
& Baseline & 56.60 $\pm$ 3.10 & 64.77 $\pm$ 1.90 & 77.33 $\pm$ 2.51 & 84.86 $\pm$ 1.16 & 88.33 $\pm$ 1.04 & 92.87 $\pm$ 0.67 \\
& Baseline LS & 60.44 $\pm$ 2.05 & 66.41 $\pm$ 2.14 & 80.54 $\pm$ 1.94 & 86.98 $\pm$ 0.99 & 91.12 $\pm$ 0.79 & 95.56 $\pm$ 0.18 \\
& LD RR & 71.34 $\pm$ 2.19 & 73.34 $\pm$ 1.18 & 84.66 $\pm$ 0.89 & 88.30 $\pm$ 0.46 & 88.91 $\pm$ 0.36 & 89.73 $\pm$ 0.39 \\
& Baseline RR & 59.20 $\pm$ 2.18 & 65.22 $\pm$ 2.29 & 77.34 $\pm$ 1.68 & 84.70 $\pm$ 0.81 & 87.87 $\pm$ 0.68 & 92.39 $\pm$ 0.53 \\
& Baseline RR LS & 60.63 $\pm$ 1.64 & 65.61 $\pm$ 1.08 & 77.39 $\pm$ 1.67 & 85.63 $\pm$ 0.95 & 88.89 $\pm$ 0.88 & 94.33 $\pm$ 0.44 \\
& DD  &  &  & & 79.5\phantom{0} $\pm$ 8.1\phantom{0} & & \\
& SLDD &  &  & & 82.7\phantom{0} $\pm$ 2.8\phantom{0} & & \\
\midrule
\parbox[t]{2mm}{\multirow{6}{*}{\rotatebox[origin=c]{90}{CIFAR-10}}} &  LD & 26.65 $\pm$ 0.94 & 29.07 $\pm$ 0.62 & 35.03 $\pm$ 0.48 & 38.17 $\pm$ 0.36 & 42.12 $\pm$ 0.56 & 41.90 $\pm$ 0.28 \\
& Baseline & 17.57 $\pm$ 1.63 & 21.66 $\pm$ 0.91 & 23.59 $\pm$ 0.80 & 27.79 $\pm$ 1.01 & 33.49 $\pm$ 0.77 & 40.44 $\pm$ 1.33 \\
& Baseline LS & 18.57 $\pm$ 0.68 & 22.91 $\pm$ 0.70 & 24.57 $\pm$ 0.83 & 29.27 $\pm$ 0.85 & 34.83 $\pm$ 0.75 & 40.15 $\pm$ 0.66 \\
& LD RR & 25.08 $\pm$ 0.39 & 28.17 $\pm$ 0.34 & 34.43 $\pm$ 0.38 & 37.59 $\pm$ 1.68 & 42.48 $\pm$ 0.25 & 44.81 $\pm$ 0.26 \\
& Baseline RR & 18.42 $\pm$ 0.59 & 21.00 $\pm$ 0.73 & 22.45 $\pm$ 0.49 & 24.46 $\pm$ 1.67 & 30.96 $\pm$ 0.49 & 39.17 $\pm$ 0.47 \\
& Baseline RR LS & 18.22 $\pm$ 0.67 & 22.31 $\pm$ 1.01 & 22.27 $\pm$ 0.75 & 24.84 $\pm$ 2.89 & 30.74 $\pm$ 0.80 & 38.86 $\pm$ 0.88 \\
& DD &  &  &  & 36.8\phantom{0} $\pm$ 1.2\phantom{0} & & \\
& SLDD &  &  &  & 39.8\phantom{0} $\pm$ 0.8\phantom{0} & & \\
\bottomrule
\end{tabular}}
\end{table*}

\begin{table*}[h!]
\caption{Dependence on real training set size. Around 5000 examples ($\approx 10$\% of all data) is sufficient. Similarly as before, we used 100 base examples. Using all examples means using 50000 examples.}
\label{tab:targetsetsizevariable}
\centering
\resizebox{1.0\textwidth}{!}{
\begin{tabular}{lcccccccc}
\toprule
Target examples & 100 & 500 & 1000 & 5000 & 10000 & 20000 & All \\
\midrule
E $\rightarrow$ M (LD) & 50.70 $\pm$ 2.33 & 61.92 $\pm$ 3.62 & 57.39 $\pm$ 4.58 & 75.44 $\pm$ 1.60 & 76.79 $\pm$ 1.12 & 77.27 $\pm$ 1.25 & 77.09 $\pm$ 1.66 \\
E $\rightarrow$ M (LD RR) & 60.67 $\pm$ 3.17 & 72.09 $\pm$ 2.40 & 65.71 $\pm$ 3.77 & 76.83 $\pm$ 2.33 & 80.66 $\pm$ 1.97 & 82.44 $\pm$ 1.64 & 82.70 $\pm$ 1.33 \\
\bottomrule
\end{tabular}}
\end{table*}

\begin{table*}[h!]
\caption{Sensitivity to number of training steps at meta-testing. We re-train the model with different numbers of steps than estimated during meta-training. The results show our method is relatively insensitive to the number of steps. The default number of steps $T_i$ (+ 0 column) was estimated as 278 for MNIST (LD), 217 for MNIST (LD RR), 364 steps for E $\rightarrow$ M (LD) and 311 steps for E $\rightarrow$ M (LD RR). Scenario with 100 base examples is reported.}
\label{tab:numstepsanalysis}
\centering
\resizebox{1.0\textwidth}{!}{
\begin{tabular}{lcccccccc}
\toprule
Steps deviation & - 50 & - 20 & - 10 & + 0 & + 10 & + 20 & + 50 & + 100 \\
\midrule
MNIST (LD) & 86.67 $\pm$ 0.51 & 86.91 $\pm$ 0.49 & 86.88 $\pm$ 0.50 & 86.77 $\pm$ 0.77 & 86.54 $\pm$ 0.68 & 87.05 $\pm$ 0.67 & 86.98 $\pm$ 0.70 & 86.59 $\pm$ 0.63 \\
MNIST (LD RR) & 88.21 $\pm$ 0.50 & 88.03 $\pm$ 0.51 & 88.32 $\pm$ 0.50 & 88.07 $\pm$ 0.46 & 88.10 $\pm$ 0.38 & 87.95 $\pm$ 0.45 & 87.98 $\pm$ 0.45 & 87.74 $\pm$ 0.62 \\
E $\rightarrow$ M (LD) & 77.28 $\pm$ 1.01 & 76.88 $\pm$ 1.97 & 77.26 $\pm$ 1.92 & 77.10 $\pm$ 1.74 & 76.40 $\pm$ 2.37 & 76.76 $\pm$ 2.18 & 77.80 $\pm$ 1.49 & 77.81 $\pm$ 1.23 \\
E $\rightarrow$ M (LD RR) & 81.84 $\pm$ 1.75 & 81.64 $\pm$ 1.76 & 81.45 $\pm$ 2.01 & 81.46 $\pm$ 1.70 & 81.55 $\pm$ 1.80 & 80.96 $\pm$ 1.74 & 81.34 $\pm$ 1.64 & 80.73 $\pm$ 2.45 \\
\bottomrule
\end{tabular}}
\end{table*}

\begin{table*}[h!]
\caption{Sensitivity of original DD to number of steps. DD is very sensitive to using the specific number of steps. We take the first N steps, keep their original learning rates, and assign learning rates of 0 to the remaining steps. When we do 5 more steps than the original (30), we perform the final 5 steps with an average learning rate.}
\label{tab:origddstepssensitivity}
\centering
\resizebox{1.0\textwidth}{!}{
\begin{tabular}{lcccccc}
\toprule
Steps & 10 & 15 & 20 & 25 & 30 & 35 \\
\midrule
MNIST & 35.65 $\pm$ 11.19 & 43.25 $\pm$ 11.47 & 54.85 $\pm$ 12.28 & 52.54 $\pm$ 11.40 & 77.32 $\pm$ \phantom{0}5.08 & 53.89 $\pm$ 10.51 \\
CIFAR-10 & 21.14 $\pm$ \phantom{0}4.08 & 22.50 $\pm$ \phantom{0}4.24 & 27.08 $\pm$ \phantom{0}3.22 &  28.54 $\pm$ \phantom{0}2.41 & 35.20 $\pm$ \phantom{0}1.09 & 28.80 $\pm$ \phantom{0}3.33\\
\bottomrule
\end{tabular}}
\end{table*}

\begin{table*}[h!]
\caption{Sensitivity of original DD to learning rates. DD is sensitive to using the specific learning rates.}
\label{tab:origddlrsensitivity}
\centering
\begin{tabular}{lcc}
\toprule
Learning rate & Optimized & Average of optimized \\
\midrule
MNIST & 77.32 $\pm$ 5.08 & 62.38 $\pm$ 13.11 \\
CIFAR-10 & 35.20 $\pm$ 1.09 & 30.59 $\pm$ \phantom{0}3.95 \\
\bottomrule
\end{tabular}
\end{table*}

\begin{table*}[h!]
\caption{Sensitivity of original DD to order of examples. DD is sensitive to using the specific order of training examples.}
\label{tab:origddordersensitivity}
\centering
\begin{tabular}{lccc}
\toprule
Order & Original & Shuffled within epoch & Shuffled across epochs \\
\midrule
MNIST & 77.32 $\pm$ 5.08 &  50.20 $\pm$ 12.83 & 62.67 $\pm$ 10.82 \\
CIFAR-10 & 35.20 $\pm$ 1.09 & 24.65 $\pm$ 2.16 & 22.59 $\pm$ \phantom{0}3.23 \\
\bottomrule
\end{tabular}
\end{table*}

\begin{table*}[h!]
\caption{Transferability of distilled labels across different architectures (second-order method). The upper part of the table shows performance of various test models when trained on distilled labels synthetised with AlexNet only. The middle part shows the baseline performance of training models with different architectures on true labels. The lower part shows that distilled labels work even in cross-dataset scenario (labels trained with AlexNet only). The results clearly suggest the distilled labels generalize across different architectures.}
\label{tab:archor2}
\centering
\resizebox{1.0\textwidth}{!}{
\begin{tabular}{lcccccc}
\toprule
Base examples & 10 & 20 & 50 & 100 & 200 & 500 \\
\midrule
CIFAR-10 LD \\
AlexNet & 26.09 $\pm$ 0.58 & 30.41 $\pm$ 0.81 & 35.21 $\pm$ 0.50 & 38.39 $\pm$ 0.62 & 40.98 $\pm$ 0.50 & 42.78 $\pm$ 0.29 \\
LeNet & 19.33 $\pm$ 2.50 & 24.17 $\pm$ 1.42 & 28.14 $\pm$ 1.37 & 32.65 $\pm$ 1.22 & 36.60 $\pm$ 1.46 & 39.67 $\pm$ 0.85 \\
ResNet-18 & 17.97 $\pm$ 1.23 & 24.64 $\pm$ 0.92 & 27.36 $\pm$ 1.07 & 31.01 $\pm$ 0.84 & 35.33 $\pm$ 0.97 & 39.32 $\pm$ 0.61 \\
\midrule
CIFAR-10 baseline \\
AlexNet & 14.35 $\pm$ 1.39 & 16.72 $\pm$ 0.76 & 21.08 $\pm$ 0.93 & 25.39 $\pm$ 0.86 & 31.39 $\pm$ 1.12 & 37.17 $\pm$ 1.58 \\
LeNet & 13.20 $\pm$ 1.86 & 15.31 $\pm$ 1.09 & 18.15 $\pm$ 0.82 & 21.63 $\pm$ 1.45 & 25.87 $\pm$ 1.26 & 32.99 $\pm$ 0.86 \\
ResNet-18 & 13.80 $\pm$ 1.19 & 18.29 $\pm$ 1.43 & 20.56 $\pm$ 0.75 & 23.44 $\pm$ 1.04 & 28.98 $\pm$ 1.05 & 33.16 $\pm$ 1.12 \\
\midrule
CUB to CIFAR-10 LD \\
AlexNet & 25.95 $\pm$ 0.90 & 27.73 $\pm$ 1.09 & 31.00 $\pm$ 0.78 & 34.99 $\pm$ 0.69 & 37.83 $\pm$ 0.65 & 39.44 $\pm$ 0.53 \\
LeNet & 20.18 $\pm$ 1.56 & 23.15 $\pm$ 1.72 & 26.16 $\pm$ 1.55 & 28.73 $\pm$ 1.44 & 30.71 $\pm$ 1.80 & 35.41 $\pm$ 0.88 \\
ResNet-18 & 17.12 $\pm$ 1.32 & 17.80 $\pm$ 1.26 & 21.22 $\pm$ 1.04 & 23.38 $\pm$ 0.95 & 23.39 $\pm$ 0.90 & 26.71 $\pm$ 0.89 \\
\bottomrule
\end{tabular}}
\end{table*}

\begin{table*}[h!]
\caption{Transferability of distilled labels across different architectures (RR method). The upper part of the table shows performance of various test models when trained on distilled labels synthetised with AlexNet only. The middle part shows the baseline performance of training models with different architectures on true labels. The lower part shows that distilled labels work even in cross-dataset scenario (labels trained with AlexNet only). The results clearly suggest the distilled labels generalize across different architectures. Note that lower RR results for ResNet-18 may be caused by significantly lower dimensionality of the RR layer (64 features + 1 for bias), while AlexNet and LeNet have 192 features + 1 for bias in the RR layer.}
\label{tab:archrr}
\centering
\resizebox{1.0\textwidth}{!}{
\begin{tabular}{lcccccc}
\toprule
Base examples & 10 & 20 & 50 & 100 & 200 & 500 \\
\midrule
CIFAR-10 LD \\
AlexNet & 26.78 $\pm$ 0.84 & 29.51 $\pm$ 0.41 & 34.71 $\pm$ 0.45 & 38.29 $\pm$ 0.92 & 41.14 $\pm$ 0.37 & 42.71 $\pm$ 0.27 \\
LeNet & 20.24 $\pm$ 2.06 & 23.58 $\pm$ 1.62 & 28.05 $\pm$ 1.66 & 30.64 $\pm$ 1.47 & 34.52 $\pm$ 1.16 & 38.75 $\pm$ 0.96 \\
ResNet-18 & 16.63 $\pm$ 0.88 & 17.98 $\pm$ 1.38 & 23.03 $\pm$ 1.21 & 26.66 $\pm$ 0.73 & 31.08 $\pm$ 0.92 & 36.37 $\pm$ 0.81 \\
\midrule
CIFAR-10 baseline \\
AlexNet & 13.37 $\pm$ 0.75 & 17.20 $\pm$ 0.50 & 19.07 $\pm$ 0.75 & 24.72 $\pm$ 0.53 & 29.94 $\pm$ 0.65 & 36.20 $\pm$ 0.97 \\
LeNet & 12.33 $\pm$ 0.88 & 14.28 $\pm$ 0.74 & 17.31 $\pm$ 0.97 & 20.61 $\pm$ 1.13 & 24.15 $\pm$ 0.98 & 28.92 $\pm$ 0.80 \\
ResNet-18 & 14.04 $\pm$ 1.20 & 16.60 $\pm$ 1.25 & 18.75 $\pm$ 1.17 & 22.61 $\pm$ 1.03 & 28.03 $\pm$ 0.72 & 33.72 $\pm$ 1.74 \\
\midrule
CUB to CIFAR-10 LD \\
AlexNet & 26.08 $\pm$ 1.14 & 29.37 $\pm$ 0.36 & 31.46 $\pm$ 3.94 & 35.74 $\pm$ 0.81 & 37.26 $\pm$ 1.63 & 40.94 $\pm$ 4.61 \\
LeNet & 22.69 $\pm$ 2.09 & 24.42 $\pm$ 1.53 & 26.35 $\pm$ 1.26 & 29.84 $\pm$ 1.36 & 31.68 $\pm$ 1.09 & 35.66 $\pm$ 1.98 \\
ResNet-18 & 16.23 $\pm$ 1.44 & 17.44 $\pm$ 0.84 & 20.06 $\pm$ 0.74 & 23.48 $\pm$ 1.00 & 21.31 $\pm$ 0.82 & 29.86 $\pm$ 0.93 \\
\bottomrule
\end{tabular}}
\end{table*}

\begin{table*}[h!]
\caption{Sensitivity of original DD to a change in architecture (trained with AlexNet). The same order of examples used as during training, with the optimized learning rates. We have not been able to easily integrate ResNet-18 to the implementation provided by the authors \citep{Wang2018DatasetDistillation}.}
\label{tab:diffarchdd}
\centering
\begin{tabular}{lccc}
\toprule
 & AlexNet & LeNet \\
\midrule
CIFAR-10 & 35.20 $\pm$ 1.09 & 25.92 $\pm$ 2.35 \\
\bottomrule
\end{tabular}
\end{table*}

\begin{table*}[h!]
\caption{Transferability of RR synthetic labels to standard model training (both within-dataset and cross-dataset scenarios evaluated). 100 base examples used. Datasets: E = EMNIST, M = MNIST, K= KMNIST, B = CUB, C = CIFAR-10, K-49 = Kuzushiji-49.}
\label{tab:rrtransferability}
\centering
\begin{tabular}{lcc}
\toprule
 & RR model & Standard model \\
\midrule
MNIST & 88.25 $\pm$ 0.37 & 87.07 $\pm$ 0.64  \\
CIFAR-10 & 38.40 $\pm$ 0.41 & 37.16 $\pm$ 0.59  \\
\midrule
CIFAR-100 & 10.96 $\pm$ 1.08 & \phantom{0}8.93 $\pm$ 0.27  \\
\midrule
E $\rightarrow$ M & 80.09 $\pm$ 1.84 & 76.04 $\pm$ 2.29  \\
E $\rightarrow$ K & 58.14 $\pm$ 0.91 & 50.37 $\pm$ 2.75  \\
B $\rightarrow$ C & 34.81 $\pm$ 6.46 & 35.76 $\pm$ 0.54  \\
E $\rightarrow$ K-49 & 17.56 $\pm$ 1.62 & 13.28 $\pm$ 1.62  \\
\bottomrule
\end{tabular}
\end{table*}

\begin{table*}[h!]
\caption{Dataset distillation of images. The results show we have not obtained strong and stable results when distilling synthetic images rather than labels (likely because of the complexity of the flexible task). It may be possible to obtain better results on distilling images with our approach, but it likely requires a lot more tuning than we have done.}
\label{tab:ddimagesanalysis}
\centering
\begin{tabular}{lcccc}
\toprule
Base examples & 10 & 20 & 50 & 100 \\
\midrule
MNIST (DD) & 55.48 $\pm$ 4.42 & 17.90 $\pm$ 4.54 & 34.40 $\pm$ 4.49 & 34.44 $\pm$ 6.15 \\
MNIST (DD RR) & 22.33 $\pm$ 2.75 & 49.30 $\pm$ 2.72 & 30.16 $\pm$ 5.32 & 36.43 $\pm$ 5.50 \\
CIFAR-10 (DD) & 12.99 $\pm$ 1.41 & 12.09 $\pm$ 0.99 & 16.01 $\pm$ 1.48 & 17.92 $\pm$ 1.71 \\
CIFAR-10 (DD RR) & 20.20 $\pm$ 0.38 & 22.71 $\pm$ 0.79 & 26.04 $\pm$ 1.22 & 27.05 $\pm$ 1.86 \\
\bottomrule
\end{tabular}
\end{table*}

\begin{figure}[h!]
    \centering
    \includegraphics[width = \textwidth]{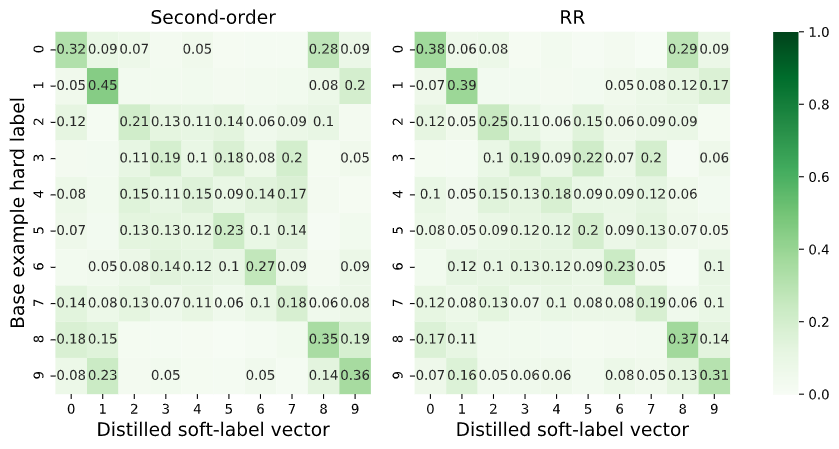}
    \caption{Distribution of label values across base example hard labels and distilled soft-label vectors. Within-dataset CIFAR-10 scenario with 100 base examples is shown. We can see that to a certain extent the original classes are recovered, but a lot of non-trivial information is added that presumably leads to strong improvements over a baseline with true or smooth labels. Numbers are shown when the values are at least 0.05.}
\label{fig:labelsanalysisc10}
\end{figure}

\begin{figure}[h!]
    \centering
    \includegraphics[width = \textwidth]{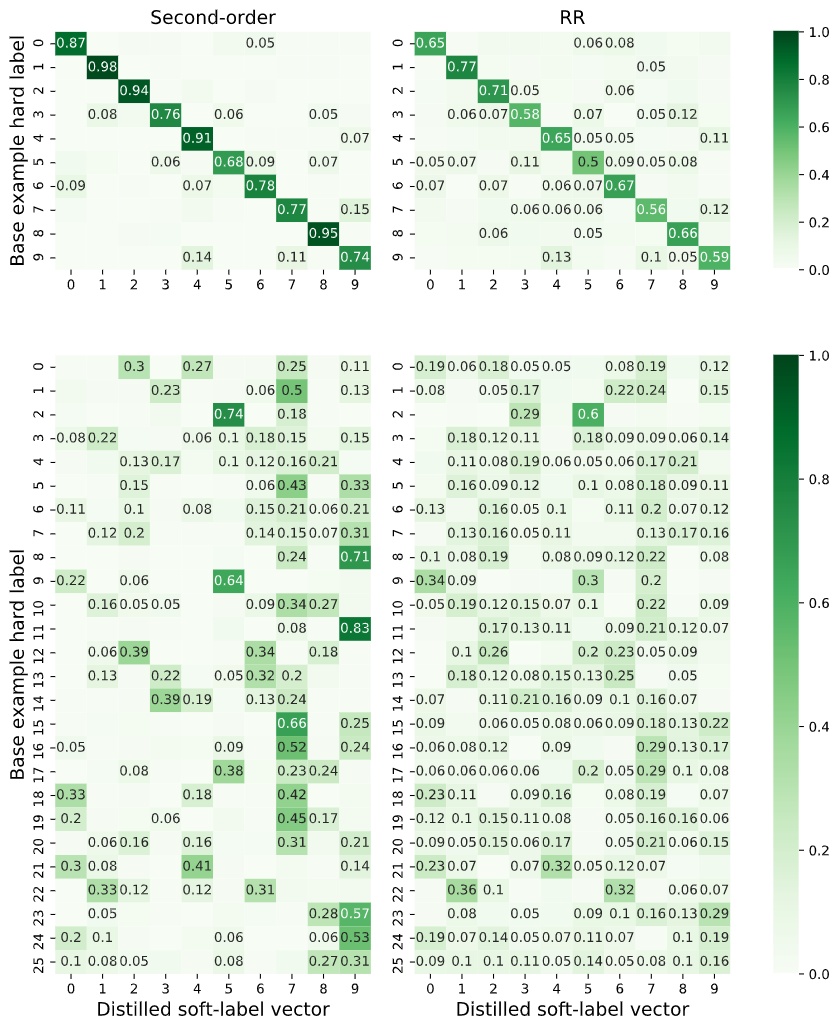}
    \caption{Distribution of label values across base example hard labels and distilled soft-label vectors. The upper row shows the mean distilled labels for different original classes for within-dataset MNIST scenario with 100 base examples. We can see that to a large extent the original labels are preserved with an additional noise on visually similar classes such as 4 and 9. At the same time, some non-trivial information is learned, especially for our RR method. The lower row shows the mean distilled labels for different original EMNIST (``English'') classes used to recognize KMNIST (``Japanese'') characters. 100 base examples scenario. Numbers are shown when the values are at least 0.05.}
\label{fig:labelsanalysis}
\end{figure}

\begin{figure}[t]
\begin{center}
 \centering
 {\includegraphics[width=\columnwidth]{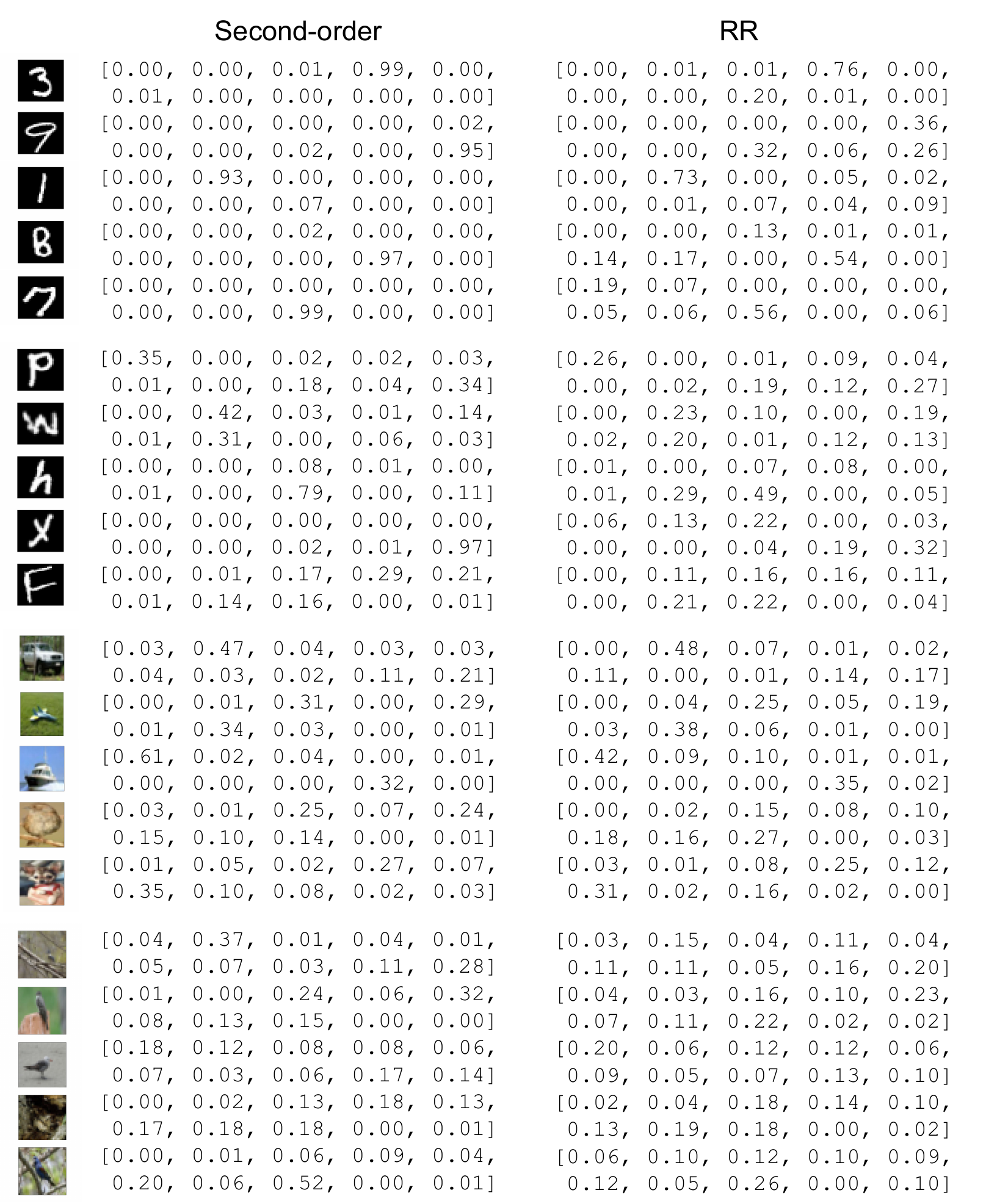}}
\vspace{-0.3cm}
\caption{Examples of distilled labels for both second-order and RR label distillation. Scenarios: 1) within-dataset MNIST, 2) cross-dataset EMNIST (``English'') source to KMNIST (``Japanese'') target, 3) within-dataset CIFAR-10, 4) cross-dataset CUB (birds) source to CIFAR-10 target. Five base examples from each scenario are shown in the order described.}
\label{fig:examplelabels}
\end{center}
\vskip -5mm
\end{figure}

\end{document}